\documentclass[10pt, conference, letterpaper]{IEEEtran}
\IEEEoverridecommandlockouts
\usepackage{diagbox}
\usepackage{subfig}
\usepackage{setspace}
\usepackage{float}
\usepackage{verbatim}
\usepackage{amsmath}
\usepackage{amssymb}
\usepackage{multirow}
\usepackage{algorithm}
\usepackage{algorithmic}
\usepackage{arydshln}
\usepackage{caption} 
\usepackage{lipsum}
\usepackage{booktabs}
\usepackage{color}
\usepackage{amsfonts}

\usepackage{array,hhline}
\usepackage{wasysym}
\usepackage{graphicx}

\usepackage{tikz}
\DeclareRobustCommand*{\circled}[1]{\lower.7ex\hbox{\tikz\draw (0pt, 0pt)%
    circle (.5em) node {\makebox[1em][c]{\small #1}};}}
\usepackage{threeparttable}
\usepackage{caption}
\newcommand{\white}[1]{ {\textcolor{white} {#1}} }

\def\name{\text{TaintRadar}}
\def\patch{\text{adversarial objects}}
\def\glassesattack{\text{accessory attack}}
\def\patchattack{\text{adversarial patch}}

\definecolor{mypink1}{rgb}{0.858, 0.188, 0.478}

\begin{document}

%\title{\Large \bf 
% \title{\fontsize{23}{25}\selectfont Detecting Localized Adversarial Examples: A Generic Approach using Critical Region Analysis}
\title{Detecting Localized Adversarial Examples: A Generic Approach using Critical Region Analysis}

% \author{
% {\rm Anonymous Author(s)}
% \\
% Your Institution
% \and
% {\rm Second Name}\\
% Second Institution
% copy the following lines to add more authors
% \and
% {\rm Name}\\
% %Name Institution
% } % end author
\author{
  \IEEEauthorblockN{
    Fengting Li\IEEEauthorrefmark{1}\IEEEauthorrefmark{2},
    Xuankai Liu\IEEEauthorrefmark{1}\IEEEauthorrefmark{2},
    Xiaoli Zhang\IEEEauthorrefmark{1}\IEEEauthorrefmark{2}\IEEEauthorrefmark{3},
    Qi Li\IEEEauthorrefmark{1}\IEEEauthorrefmark{2},
    Kun Sun\IEEEauthorrefmark{4},
    and Kang Li\IEEEauthorrefmark{5}
  }
  \IEEEauthorblockA{
    \IEEEauthorrefmark{1}Institute for Network Sciences and Cyberspace \& Department of Computer Science and Technology, Tsinghua University\\     
    \IEEEauthorrefmark{2}Beijing National Research Center for Information Science and Technology (BNRist), Tsinghua University\\
    \IEEEauthorrefmark{3}Gemini Lab, Alibaba Group
    \IEEEauthorrefmark{4}Department of Information Sciences and Technology, CSIS, George Mason University 
    \IEEEauthorrefmark{5}Baidu
  }
  \IEEEauthorblockA{
    \{lft18, liuxk18\}@mails.tsinghua.edu.cn, xiaoli.z@outlook.com, qli01@tsinghua.edu.cn, ksun3@gmu.edu, kangli.ctf@gmail.com
  }
  
% {\phantom{.}}
}

\maketitle

\begin{abstract}
Deep neural networks (DNNs) have been applied in a wide range of applications, e.g., face recognition and image classification;  however, they are vulnerable to adversarial examples. By adding a small amount of imperceptible perturbations, an attacker can easily manipulate the outputs of a DNN. Particularly, the localized adversarial examples only perturb a small and contiguous region of the target object, so that they are robust and effective in both digital and physical worlds.  Although the localized adversarial examples have more severe real-world impacts than traditional pixel attacks, they have not been well addressed in the literature. In this paper, we propose a generic defense system called TaintRadar to accurately detect localized adversarial examples via analyzing critical regions that have been manipulated by attackers. The main idea is that when removing critical regions from input images, the ranking changes of adversarial labels will be larger than those of benign labels. Compared with existing defense solutions, TaintRadar can effectively capture sophisticated localized partial attacks, e.g., the eye-glasses attack, while not requiring additional training or fine-tuning of the original model's structure. Comprehensive experiments have been conducted in both digital and physical worlds to verify the effectiveness and robustness of our defense.

% Deep neural networks (DNNs) have been applied in a wide range of applications, e.g., face recognition and image classification;  however, they are vulnerable to adversarial examples. By adding a small amount of imperceptible perturbations, an attacker can easily manipulate the outputs of a DNN. Particularly, the localized adversarial examples only perturb a small and contiguous region of the target object, so that they are robust and effective in both digital and physical worlds.  Though the localized adversarial examples have more severe real-world impacts than traditional pixel attacks, they have not been well addressed in the literature. In this paper, we propose a generic defense system called TaintRadar to accurately detect localized adversarial examples via analyzing critical regions that have been manipulated by attackers. The main idea is that when removing critical regions from input images, the ranking changes of adversarial labels will be larger than those of benign labels. In particular, TaintRadar can effectively capture sophisticated localized partial attacks, e.g., the eye-glasses attack.  
% Compared with existing defense solutions, TaintRadar does not require additional training or fine-tuning of the original model's structures.
% Comprehensive experiments have been conducted in both digital and physical worlds to verify the effectiveness and robustness of our defense. 
\end{abstract}

%!TEX root=./main.tex
\section{Introduction}
\label{sec:intro}

With the rapid development of deep learning techniques, neural networks have been applied in various applications, e.g., face recognition~\cite{parkhi2015deep} and object detection~\cite{redmon2016you}, to achieve a better accuracy than traditional machine learning methods. However, neural networks are vulnerable to adversarial examples~\cite{szegedy2013intriguing}, where attackers can generate perturbations against neural network models to raise misclassification. For example, an adversarial attack can generate imperceptible pixel-level perturbations  in an image so that the image looks the same as the original one for humans, but  easily deceives the classifier into generating an entirely different label~\cite{moosavi2016deepfool,moosavi2017universal,dong2019evading,carlini2017towards, su2019one}. A number of defenses have been proposed to detect such pixel-level attacks~\cite{dziugaite2016study,kurakin2016adversarial,xu2017feature,papernot2016distillation}. One limitation of this imperceptible perturbation is that it can hardly be applied in the real world due to the difficulty on manipulating the images of the real world in the granularity of pixels. 

As one type of adversarial examples, the localized adversarial examples focus on perturbing a small and contiguous region that is visible to human eyes and can be printed out to launch attacks in the real world. By adding physical penalty constraints, localized adversarial examples become robust to various physical conditions such as locations, sizes, and even different background patterns. Localized adversarial examples can be classified into two categories, namely, {\em localized universal attacks} and {\em localized partial attacks}. The localized universal attacks aim to raise false predictions on arbitrary inputs by using one adversarial object. For example, adversarial patches~\cite{brown2017adversarial} can be printed to perform generalized attacks when it appears in the scene and is robust under different real-world environments. In contrast, the localized partial attacks focus on manipulating the predictions on one or a small subset of labels. For example, an adversary can easily fool the face recognition system by wearing a decoration such as a pair of glasses~\cite{sharif2016accessorize}.  
%signficant damage to the security in many fields, e.g., surveillance. They are more harmful compared with imperceptible noise attacks.

Though it is well known that localized adversarial examples can cause severe threats in the real world, a generic and  deployable defense system is still missing. %Most of existing defenses, e.g., feature squeezing~\cite{xu2017feature} and JPEG compression~\cite{dziugaite2016study}, only aimed at imperceptible noise attacks, which cannot effectively detect localized attacks. 
In general, model robustness approaches~\cite{chiang2020certified, wu2019defending} require extra training with high training overhead and are  difficult to work at ImageNet scale~\cite{chiang2020certified, kurakin2016adversarial}. Also, they cannot be applied to guard existing models. Several defense solutions have been proposed to defeat localized adversarial examples~\cite{hayes2018visible, naseer2019local,chou2018sentinet}; however, as shown in Table~\ref{tab:Relatedwork}, they all have significant limitations, 
% \ft{ Chiang et al.~\cite{chiang2020certified} borrows the idea of interval bound propagation~\cite{gowal2018effectiveness, mirman2018differentiable} to address the problem of localized adversarial examples, which gives a certificate when an output lies in an interval bound introduced during training, producing a lower bound on adversarial accuracy. Adversarial training~\cite{madry2017towards} is also applied to augment the robustness of a model under the threat of localized adversarial examples~\cite{wu2019defending}. However, }
Hayes et al.~\cite{hayes2018visible} propose a digital watermarking (DW) mechanism based on the observation that the density of salient pixels with respect to the final output is larger in \patch{} than that in benign objects. %It further mitigates the adversarial effects by replacing the pixels inside the dense region with averaged values of nearby pixels. 
Based on a similar observation that pixel values change more drastically inside the perturbed area, Local Gradient Smoothing  (LGS) mechanism~\cite{naseer2019local} can mitigate the adversarial effects of perturbations by smoothing the gradient inside that area. These two input transformation approaches are vulnerable under the threat of adaptive attacks~\cite{chiang2020certified}. Also,  LGS is sensitive to the sizes and shapes of patches and not attack-agnostic. 
SentiNet~\cite{chou2018sentinet} can detect adversarial objects by testing the behavior of \patch{} with benign test samples~\cite{selvaraju2017grad}; however, its detection effectiveness heavily relies on the generalization of the \patch{}. Overall, none of  existing defenses can detect the stealthy partial attacks including eye-glasses attacks~\cite{sharif2016accessorize}. 
%chooses to trade the generalizability with the stealthiness}, none of existing defenses can detect those partial attacks including eye-glasses attacks~\cite{sharif2016accessorize}.  

\begin{table*}[t]
    \vspace{-0.15in}
    \caption{Comparison with existing defenses against \patch{}. 
    $\large\Circle$ denotes that the attack cannot be detected;   
    $\large\CIRCLE$ denotes that the defense  can defeat the attacks;  $\checkmark$/$\times$ illustrates whether the defense can achieve the corresponding property. 
    % and $\large\LEFTcircle$ illustrates that some universal attacks can be detected.
    }
    \label{tab:Relatedwork}
    \vspace{-0.1in}
    \begin{threeparttable}
        \centering
        \resizebox{0.9\textwidth}{!}{
        \begin{tabular}{c|c|c|c|c|c|c|c}
            \toprule[2pt]
            & \multicolumn{2}{c|}{Localized Adversarial Example}          & \multicolumn{5}{c}{ Desired Properties of Defenses}                              \\ 
            \hline
            Defenses  & Partial Attack                 & Universal Attack              &  Genericity              & Attack-agnostic                & Agility    & Lightweight &Robustness     \\ 
            \hline

            LGS~\cite{naseer2019local}  &         	
            $\large\Circle$    & $\large\Circle$    & 
            $\times$ &  $\times$ &  $\checkmark$ & $\checkmark$ &   $\times$    \\ 
            \hline

            DW~\cite{hayes2018visible}  & 
            $\large\Circle$     & $\large\Circle$ 
            &$\times$ &$\checkmark$ &$\checkmark$ & $\times$ &  $\times$            \\ 
            \hline

            SentiNet~\cite{chou2018sentinet} &
            $\large\Circle$   & $\large\CIRCLE$           
            &$\times$ &$\checkmark$ &$\checkmark$  & $\times$ &    $\checkmark$      \\ 
            \hline
            \textbf{\name{}}        &
            $\large\CIRCLE$      &$\large\CIRCLE$ 
            &$\checkmark$  &$\checkmark$ & $\checkmark$ & $\checkmark$  & $\checkmark$          \\ 
%         %\hline
            \bottomrule[2pt]
        \end{tabular}
        }
    %  \begin{tablenotes}
    %      \footnotesize
    %      \item[1]Februus is not designed for adversarial example detection; however, the mechanism can be generalized to detect localized adversarial examples.
     
    %  \end{tablenotes}

% %\multicolumn{8}{l}{$^1$ Note that, Februus is not designed for adversarial example detection. However, the mechanism can be generlized to detect localized adversarial examples.}\\
    \end{threeparttable}
    \vspace{-0.15in}
\end{table*}

In this paper, we develop \name{}, a generic defense system that can detect different types of localized adversarial examples in both digital world and physical world scenarios. \name{} accurately detects attacks by identifying critical regions that have been manipulated, where a critical region is the region that supports the final prediction. When removing a critical region, the ranking of the predicted label will be lowered along with a corresponding probability. However, after removing critical regions of the same small sizes from benign and adversarial images, the ranking changes of predicted labels on adversarial images will be larger than benign images. To accurately detect localized adversarial examples, \name{} first estimates the critical regions of input images and then measures the ranking changes of the input images before and after removing the critical regions.
%\ftdel{In particular, we utilize the top-K labels of the corresponding regions to %measure the ranking changes in benign and malicious images 
%to enlarge the difference of ranking changes between benign and malicious images 
%so that we can accurately identify the difference between benign and malicious images and detect localized adversarial attacks.}
In particular, we utilize the top-K changes of logits to refine the estimated region, which can effectively enlarge the difference between benign and malicious images and increase the accuracy on identifying the attack.

\name{} provides several critical properties on detecting localized adversarial examples. First, it is a generic detection system that can detect various localized adversarial examples including both localized universal attacks and localized partial attacks. Second, it is attack-agnostic, since it does not require any prior knowledge of the attacks. Third, it offers good agility for deployment as a plug-and-play mechanism, since it can work with various neural networks without modifying the models. Fourth, it is lightweight since it does not require training any extra model and can detect attacks in real time. Finally,  %\ftdel{it is robust by adapting to different attack variants.} 
it is robust against different attack variants including the adaptive attacks.

We systematically evaluate the performance of \name{} using two typical scenarios, i.e., scene classification and face recognition scenarios. %In particular, we reproduce attacks in the physical world and measure the effectiveness of our defense in real world. 
%We validate the effectiveness of detecting localized partial and universal attacks\footnote{Note that, since }}. 
In the scene classification scenario, \name{} can achieve more than 93\% True Positive Rate (TPR) with a 6\% False Positive Rate (FPR) when we use different attack variables to construct various partial and universal attacks. Moreover, \name{} can effectively detect partial attacks in the face recognition scenario and achieve more than 73\% TPR in detecting partial attacks. In contrast,  SentiNet~\cite{chou2018sentinet} can only achieve a 0.7\% TPR on the same task. In addition, we can achieve similar detection performance in the physical environment. Furthermore, we validate the robustness of \name{} under various advanced attack settings. The experimental results confirm that \name{} can achieve comparable detection performance even under advanced attacks such as adaptive attacks.  

In summary, we make the following contributions:
\begin{itemize}

\item We propose a defense solution to defeat localized adversarial examples. 
The basic idea is to measure the difference before and after removing the critical regions that may be leveraged by the attackers. It is the first generic defense that can be applied to detect various localized adversarial examples (including partial and universal attacks) in different scenarios, without requiring any prior knowledge of the attacks. 

%\item We develop an adversarial region detection algorithm to estimate potential critical regions that may be leveraged by the attacks and then identify adversarial regions by  measuring the differences before and after removing the critical regions.  
%\item TaintRadar is based on the existing DNN model, with no need to retrain the model, which makes TaintRadar convenient to deploy. Besides, no prior knowledge is required.

\item We evaluate the effectiveness and overhead of \name{} under different attacking settings in both digital and physical worlds. The experimental results show that \name{} can effectively detect both partial attacks and universal attacks. Since it requires no extra training and only incurs around 80 $ms$ extra processing delay, it is promising to be deployed  in time-bounded real systems.

\item We conduct experiments to demonstrate the robustness of \name{} against advanced attacks with different adversarial variants, e.g., the adaptive attacks that have the white-box knowledge of \name{}.

%\item \ft{\name{} requires no extra training and incurs the overhead of only around 80ms, which makes it readily deployable in time-bounded systems.}
%. The results show that TaintRadar is effective against adversarial patches of different sizes, shapes and attack capabilities. Beside, we also demonstrate the robustness of our method by taking adaptive attacks into account.

%\item Not only in digital environment, we also conduct physical experiments in different scenarios, including scene classification and face recognition. We verify that TaintRadar can perform well in the real world.

\end{itemize}

\section{Background on Neural Networks}
A neural network consisting of multiple layers can be expressed as a function $F_\theta(x) = y$, where $\theta$ is the model parameters, $x$ is the input (e.g., an image), and $y$ is the  output (e.g., the result of face recognition). 
In this paper, we focus on convolutional neural networks (CNN) that can be used as $m$-class image classifiers. 
One CNN is mainly composed of convolutional layers, fully connected layers, and a softmax layer. A convolutional layer consists of multiple convolutional kernels, and each kernel can extract specific high-level visual features called feature maps. The fully connected layers  map these features to classification labels.  The softmax layer, as the last layer, takes the results of the prior layer (called logits $Z(x)$) as input and calculate the output vector as a probability distribution, which meets $0 \le y_i \le 1$ and $y_1+y_2+\cdots+y_m = 1$. The final classification label $l$ for the input $x$ is the one with the highest probability $y_l$. We use $r_l$ to denote the ranking of the probability corresponding to a label $y_l$ among all probabilities.

% %\subsection{Attack Model}
% \begin{figure}[h]
% 	\centering
% 	\subfloat[Scene adversarial object]{
% 		\label{fig:scenePatch}%%
% 		\begin{minipage}[t]{1.0\linewidth}
% 			\centering
% 			\includegraphics[width=1.0in]{./figs/cougar.png}
% 			%\quad \quad \quad
% 			\includegraphics[width=1.0in]{./figs/cougar_patch.png}
% 			\includegraphics[width=1.0in]{./figs/toaster.png}
% 			\\
% 			\footnotesize{%Original label: 
% 			\emph{Cougar} \quad \quad  \quad \quad \quad %Adversarial label: 
% 			\emph{Toaster} \quad \quad \quad \quad \quad \emph{Toaster}
% 			}
% 	    \end{minipage}}

% 	\subfloat[Facial adversarial object]{
% 		\label{fig:facePatch}%%
% 		\begin{minipage}[t]{1.0\linewidth}
% 			\centering
% 			\includegraphics[width=1.0in]{./figs/origin_Elaine_Hendrix.jpg}
% 			%\quad \quad  \quad
% 			\includegraphics[width=1.0in]{./figs/glass_Elaine_Hendrix.jpg}
% 			\includegraphics[width=1.0in]{./figs/A.J.png}
% 			\\
% 			\footnotesize{
% 			%Original label: 
% 			\emph{Elaine Hendrix} \quad \quad \quad %Adversarial label: 
% 			\emph{A.J. Buckley} 
% 			\quad \quad \quad 
% 			\emph{A.J. Buckley}
% 			}
% 	    \end{minipage}}

%     \caption{Examples of localized adversarial objects.
%     From left to right: original images, attacked images, target labels.}
% 	\label{fig:attackexample}
% \end{figure}

\section{Attack Model}
\label{AttackModel}
Attackers can generate various adversarial examples to fool a pre-trained CNN model. There are effective  attacks~\cite{szegedy2013intriguing,akhtar2018threat} that produce imperceptibly global perturbations for misclassification purpose; however, it is challenging to apply them in the physical world due to the difficulty of adding global pixel-level perturbations. In this paper, we focus on \emph{localized adversarial examples} that construct visible but inconspicuous perturbations in a small and contiguous region. For simplicity, we call such regions as {\em \patch{}}. The localized attacks can be more easily launched in the physical world by directly putting or wearing \patch{} in the scene. With the presence of \patch{} in an input image, 
the classification model can be hijacked to generate a completely different output, e.g., in facial biometric systems~\cite{sharif2016accessorize} or object recognition systems~\cite{brown2017adversarial}.
In this paper, we study both localized universal attacks and localized partial attacks.

\noindent \textbf{Localized Universal Attack.}
It aims to raise false predictions on arbitrary inputs using one adversarial object. The adversarial object is generated against multiple images with different source labels to achieve either targeted or untargeted attacks. Size variations and printability of \patch{} are two main factors to be considered when launching robust attacks in varying physical environments. \cite{brown2017adversarial} is an representative example of localized universal attacks against benign CNN models in scene classification scenario.

\noindent \textbf{Localized Partial Attack.} It focuses on manipulating  prediction results of a model on one or a small set of labels. For example, Sharif et al.~\cite{sharif2016accessorize} successfully generate a pair of eye-glasses as an adversarial accessory to achieve source-label-specific dodging or impersonation attacks. This type of attack is stealthier than the localized universal attacks, since they only need to compromise a small set of specific labels and those misclassification results may not be easily discovered. 
%!TEX root=./main.tex

\section{Overview of \name{}}
\label{sec:overview}
The goal of \name{} is to detect localized attacks by identifying critical regions that have been manipulated. A critical region is the region that contributes heavily to the final prediction result of neural networks. 
When removing  a critical region, the probability of the output label of original image (i.e., the predicted label) decreases and the corresponding ranking drops. Meanwhile, when the size of the removed region increases, the ranking will decrease more. 

% \ftdel{We observe that critical regions of malicious inputs, i.e., \patch{}, %\emph{(inputs with adversarial objects are called malicious inputs, instead of adversarial objects alone)}, 
% are usually limited to small regions with little image semantic information, while the critical regions of benign inputs are relatively large and distributed to multiple parts, each part having meaningful image semantic (e.g., an ear in a human face or a tail of an animal). 
% Therefore, we can identify localized attacks by checking the amount of image semantic information in critical regions. Fortunately, we can achieve it by measuring the ranking changes of predicted labels 
% % (see Algorithm~\ref{algo:regionEstimation}).
% }

\begin{figure}[ht]
	\centering 
	\vspace{-0.10in}
	\includegraphics[width=1.6in, height=1.15in]{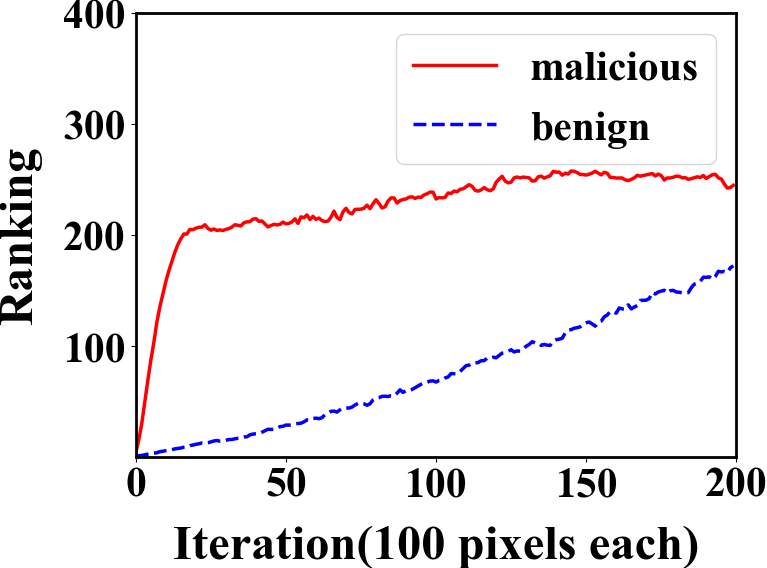}
	\caption{Ranking variations between benign and adversarial images as the sizes of removed critical regions increase.}
	\vspace{-0.15in}
	\label{fig:intuition}
\end{figure}

We conduct experiments to measure the ranking changes of predicted labels on benign and adversarial images after removing critical regions with varied sizes. The experiments consist of three steps, namely, finding critical regions, continuously removing the most important pixels from critical regions, and recording the corresponding probability ranking of original labels.
%
%\ks{We identify critical regions by leveraging Grad-CAM, a popular visual explanation tool~\cite{selvaraju2017grad}}. 
We first develop an estimation algorithm to identify the critical regions with the importance score (see Section \ref{subsec:critical}). Then, we sort the positive importance of each pixel in a descending order 
%\ftdel{Given a coarse-grained 14x14 heatmap, we resize the heatmap to the size of the original image and sort the positive importance of each pixel in a descending order.} 
%\ft{Then,} we 
and remove the 100 most important pixels each time and observe the ranking changes of the original label. Figure~\ref{fig:intuition} depicts the trend averaged over 200 images generated using the method~\cite{brown2017adversarial}, where the ranking changes of benign images with gradual region removal grow slower than those of adversarial images. Also, experiments on other attacks show a similar trend. Thus, we can have the following key observation. 

\begin{figure*}[t]
	\centering 
	\vspace{-0.10in}
	\includegraphics[width=0.85\linewidth]{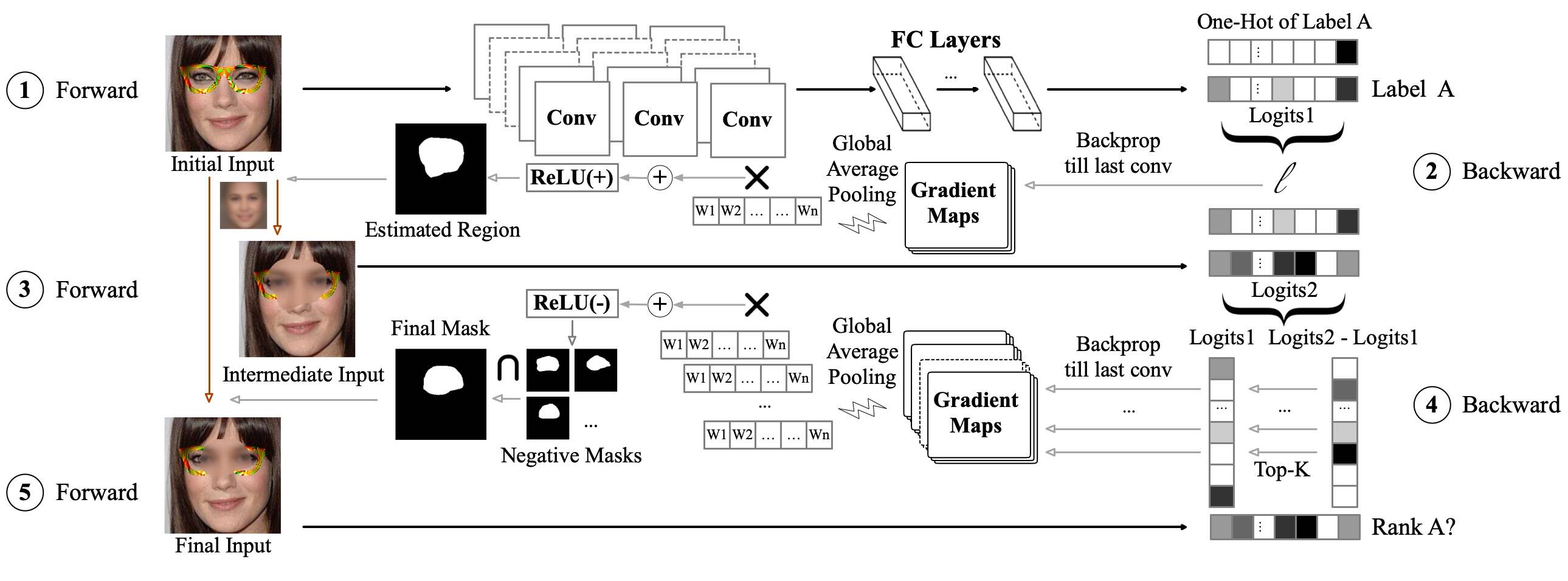}
	\vspace{-0.10in}
	\caption{Workflow of \name{}. Forward pass\circled{1}feeds the initial input to the model and gets the corresponding logits and label. We identify critical regions through backward pass\circled{2}using estimation function $\ell$ to the last convolutional layer (see Section~\ref{subsec:critical}). Then, we replace the critical regions with a filling pattern to generate an intermediate input and compute the changes of logits with forward pass\circled{3}. With the top-K logit changes, we obtain the negative masks by backward pass\circled{4}(see Section~\ref{subsec:detection}). We intersect these K negative masks to get the final mask and replace the region with the filling pattern. Finally, we use the ranking changes of the original label through forward pass\circled{5}to detect attacks.}
	\label{fig:overview}
	\vspace{-0.20in}
\end{figure*}

%\zxl{It is reasonable since the regions of benign images contain more meaningful semantic information and thus generate less impacts on the ranking than the adversarial ones.\{duplicate from previous para.?\}}

% \vspace{0.02in}
\noindent \emph{\textbf{Key Observation.} The ranking changes of predicted labels on adversarial images are larger than those of benign images after removing the same small size of critical regions from both benign and adversarial input images.}    

Figure~\ref{fig:overview} shows the high-level workflow of \name{}, which first estimates critical regions of input images and then analyzes the ranking difference before and after removing the regions for detection.
\name{} is designed as a \emph{generic} defense system that can detect both localized universal attacks and localized partial attacks in digital and physical worlds. It is \textit{attack-agnostic}, since the defense does not require any prior knowledge on attacks including attack methods or the information of adversarial objects, e.g., their shapes, sizes, and locations.  Also, \name{} is \emph{agile} to be deployed in real world scenarios since it does not need to make any changes over the neural networks. 
%or require extra training \ftadd{different with lightweight and agile defined in intro}. 
The defense is \emph{lightweight} so that it can be applied to detect attacks in real time without requiring extra training. Moreover, \name{} is \emph{robust} to effectively detect various advanced attacks, e.g., adaptive attacks with  white-box knowledge of both the neural network model and the detection method.

% \vspace{0.10in}	%\vspace{1cm}
% \vspace{-0.10in}
%!TEX root=./main.tex
% \vspace{-0.12in}
\section{System Design} % of \name{}}
\label{sec:Details}
We first introduce the estimation approach to capture the critical region of an input image. Next, we analyze the influence of removing the estimated region and utilize the top increases of logits to enlarge the difference between benign and malicious images. Finally, we differentiate benign and malicious images by checking the ranking change of the predicted label.

%\vspace{-0.15in}
\subsection{Critical Region Estimation}
\label{subsec:critical}
We first locate the small and contiguous region that may be leveraged to launch an attack. As long as we can identify such region in an input image, we can check to see if they are being attacked. Inspired by~\cite{selvaraju2017grad}, we utilize the gradient information of neurons in the last convolutional layer to generate a critical region of a given input image, where the high-level semantics are better than other convolutional layers and the preserved spatial information is also richer than the fully-connected layers. Since the \patch{} usually do not block the original object, they  not only need to increase the activation value of the target label, but also degrade the activation of the original object and some other related labels. We define a cross-entropy function to effectively capture the behaviors of both promoting and suppressing the activation. 
\begin{equation} 
l = - \mathtt{CrossEntropy}(\bar{y}^{c}, p) = \sum_{i} \bar{y_{i}}^{c} log(p_{i}), 
\end{equation}
where $\bar{y}^{c}$ is the one-hot encoding of predicted label $c$, and $p$ is the softened output of the model obtained by Eq. (\ref{eq:softlabel}). Note that we add a negative sign to the cross-entropy function, indicating that we find a critical region that boosts the current identified label while suppressing the remaining labels. Intuitively, we want to find a region that offers the largest benefit to launch an untargeted attack on the current input so that the input will be misclassified. It is most likely that this region has been leveraged to construct a successful attack.

Considering that a malicious image might have high prediction confidence on a target label, $p$ may be close to $\bar{y}^{c}$. Meanwhile, the gradient value of back-propagation may be 0 according to the rule for gradient calculation. To avoid the obtained gradient being 0, we divide logits $Z$ by a parameter $T$ before performing softmax, where $T$\textgreater 1. We can get the softened output $p_{i}$ as follows:
\begin{equation}
\label{eq:softlabel}
p_{i} = \mathtt{Softmax}(\frac{Z}{T})_{i} \\
= \frac{e^{\frac{Z^{i}}{T}}}{\sum_{k} e^{\frac{Z^{k}}{T}}}, 
\end{equation}
where $p_{i}$ is the $i$-th node value in the output vector and $Z^{i}$ is the corresponding $i$-th logit. 

We then define the importance weight $\alpha_{k}$ of the $k$th feature map $A^{k}$ as the global-average-pooling over gradients of each neuron $A_{ij}^{k}$ with respect to the function $l$.  %Thus, we can obtain that: 
\begin{equation}
\begin{aligned}
\label{eq:ourWeight}
\alpha_{k} &=\frac{1}{N} \sum_{i} \sum_{j} \frac{\partial l}{\partial A_{ij}^{k}}\
=\frac{1}{N} \sum_{i} \sum_{j} \sum_{s} \frac{\partial l}{\partial Z^{s}} \cdot \frac{\partial Z^{s}}{\partial A_{ij}^{k}},\\ 
% &=\frac{1}{NT} \sum_{i} \sum_{j}\left[\left(\bar{y}_{c}-a_{c}\right) \frac{\partial Z^{c}}{\partial A_{ij}^{k}}+\sum_{s \neq c}\left(\bar{y}_{s}-a_{j}\right) \frac{\partial Z^{s}}{\partial A_{ij}^{k}}\right] \\ 
% &=\frac{1}{NT} \sum_{i} \sum_{j}\left[\left(1-a_{c}\right) \frac{\partial Z^{c}}{\partial A_{ij}^{k}}-\sum_{s \neq c} a_{s} \frac{\partial Z^{s}}{\partial A_{i j}^{k}}\right] \\
&=\frac{1-p_{c}}{T} \cdot \frac{1}{N} \sum_{i} \sum_{j} \frac{\partial Z^{c}}{\partial A_{ij}^{k}}-\frac{p_{s}}{NT} \sum_{i} \sum_{j} \sum_{s \neq c}  \frac{\partial Z^{s}}{\partial A_{ij}^{k}},
\end{aligned}
\end{equation}
where $N$ is the number of neurons in each feature map. The derivation of $\alpha_{k}$ in Eq. (\ref{eq:ourWeight}) shows that we can leverage all the information in the output vector to estimate the region. In particular, the first half of equation indicates the region that supports the predicted label, while the second half, with a negative sign, represents the regions suppressing the other labels. Then, we perform a weighted-sum over all the feature maps. By adding a ReLU function, we get a positive heatmap that represents the critical region. Ultimately, we normalize and binarize it to get $L_{est}$ using the same threshold suggested in~\cite{selvaraju2017grad}, i.e., 0.15, to obtain the final mask, where all the pixels inside and outside the estimated region are set as 1 and 0, respectively. 
\begin{equation}
\label{eq:TaintRadar}
L_{est} = \mathtt{Binarize}\left(\frac{\mathtt{ReLU}(\sum_{k} \alpha_{k}A^{k})}{\mathtt{max}(\mathtt{ReLU}(\sum_{k} \alpha_{k}A^{k}))}\right).
\end{equation}
%Due to the normalization process, 
\noindent Note we can cancel $T$ in Eq. (\ref{eq:ourWeight}) since it has influence only on $p_{c}$ and $p_{s}$, but not the  output. However, with the unchanged relative magnitude, the critical region is stable with different $T$ values after the softening process. Our system sets $T$ to 2. 

\subsection{Critical Region Based Detection} %Counterfactual Analysis}
\label{subsec:detection}
Now we aim to accurately identify adversarial perturbations by analyzing the estimated critical regions.

\noindent \textbf{Strawman Approach. }
%ZXL-2020-1-2：先说下 strawnman 方法到底是啥。一句话说明如何区分。这里没有介绍方法。
%ZXL-2020-1-2：好多词语用的不够严谨，什么是 output distribution？这里应该是观察 variation of original labels' rankings?
%Intuitively, we can observe the variations in output distribution after removing the critical regions as illustrated before in Key Observation 1. 
%Once we can find adversarial objects, the most 
A straightforward approach is to simply remove the estimated critical regions and observe the ranking change of the original label. If the ranking of the original label changes drastically, the critical regions are most likely to be \patch, meaning the image have been attacked. However, we find that the ranking changes may be large for both benign images and adversarial images, making it difficult to distinguish them.
% ZXL-2020-1-2：逻辑一步一步需要再严谨一些。什么是 impact on the image? 不准确。
Particularly, when the estimated critical region covers most of the main object in a benign image, the rankings of the original label may change dramatically if we remove the critical region (see Figure~\ref{fig:intuition}).

\noindent \textbf{Our Approach. }
% ZXL-2020-1-2：logits 不用反复强调 before the softmax。
% ZXL-2020-1-2：先high-level 说下怎么解决 strawnman solution 的缺陷的。为什么要找 top-k 的镇压区域？
% ZXL-2020-1-2: the region that suppresses other labels？分下谁镇压谁... 这里的主动被动用的不太对。
%It is evident that Strawnman Solution has a great impact on the original label’s rankings when removing the main objects in benign images. 
We can tackle this problem in the above Strawman approach by reducing the impact on the benign images. 
%
%Based on Key Observation 2, 
We find that, in order to successfully construct an attack, adversarial objects are usually constrained in small contiguous regions where all pixels are utilized together. However, for benign inputs, different regions are relatively distributed as multiple parts and each region is with different semantics, e.g., the region may be a tail or an ear of a cat.  
Moreover, \patch{} normally do not block the original object. Thus, to successfully launch an attack, \patch{} will inevitably suppress logits of a large group of labels apart from increasing the logit of the target label.
Although both benign and malicious objects suppress logits value of other labels to achieve the highest success probability, the regions suppressing each label are centralized and scattered for malicious and benign inputs, respectively. 

Therefore, to enlarge the difference between the rankings (see Figure~\ref{fig:intuition}), we select the top-$K$ labels that are suppressed most with the largest logit increases after overlaying the critical region with some filling patterns. Then, we leverage the counterfactual explanation~\cite{selvaraju2017grad} %in Grad-CAM~\cite{selvaraju2017grad} 
to find negative critical regions. First, we apply global-average-pooling on the gradients flowing from the logit $Z^l$ to the last convolutional layer to obtain the weight of each feature map (see Eq. (\ref{eq:weight})). 
 %counterfactual explaination of Grad-CAM. 
\begin{equation}
\label{eq:weight}
\alpha_{k}^{l} = \frac{1}{N} \sum_{i} \sum_{j} \frac{\partial{Z^l}}{\partial{A_{ij}^k}},  
\end{equation}
where $l$ is one of these top-$K$ labels, and $\alpha_{k}^{l}$ is the weight of the $k$-th feature map~\cite{selvaraju2017grad}. 
%calculated in Grad-CAM~\cite{selvaraju2017grad}. 
%The weight $\alpha_{k}$ is obtained by calculating the gradient of the logits $Z^c$ (before the softmax layer)
%Then, we use the following equation to find the corresponding suppressed region of each label inversely. % (Grad-CAM can be used to find both promotion and suppression regions with repect to centain class $l$):
Second, we can use equation Eq. (\ref{eq:gradcam_negative}) to find the negative critical regions of each label. 
Note that, before feeding the weighted map into the $ReLU$ function, we add a negative sign to obtain the region of a negative impact on the current logit, namely, suppressing the prediction of label $l$.
\begin{equation}
\label{eq:gradcam_negative}
L_{negative}^{l} = \mathtt{ReLU}(-\sum_{k} \alpha_{k}^{l}A^{k}).
\end{equation}
%Now we can compute the $K$ negative critical regions as follows: 
After identifying the $K$ negative critical regions, we can first binarize them using a threshold of 0.15 according to our empirical study %~\cite{selvaraju2017grad} \ft{times the largest absolute value}, 
and then intersect these $K$ masks to generate a final mask (see Eq. (\ref{eq:final})). For adversarial inputs, the intersection of these $K$ suppression regions accounts for a significant ratio of the adversarial objects. For benign inputs, the suppression regions corresponding to these $K$ labels are scattered, so the region after the intersection will be small enough, significantly reducing the impact on benign inputs.
\begin{equation}\label{eq:final}
L_{TaintRadar} = \bigcap_{l=1}^{K} \mathtt{Binarize}\left(L_{negative}^{l}\right). 
\end{equation}
As analyzed before, the rankings of original labels are sensitive to the changes if we remove critical regions in malicious images. Small modifications in attacked regions will lead to obvious changes of the prediction outputs. Therefore, we in-paint the intersection region with a filling pattern and resend it to the classifier to get the final ranking of the original predicted label. Here, we define whether an input is malicious by a threshold $\Delta R$, i.e., the ranking change of the original predicted label. % that is obtained by directly feeding the input into the model.

\section{Experiments}
\label{sec:Experiments}
% In this section, we compare \name{} on the representative localized adversarial examples (i.e., \glassesattack{} and \patchattack{}) in two typical scenarios. We validate our defense against each attack in both digital and physical worlds with multiple attack variants. To further demonstrate the robustness of \name{}, we extensively evaluate it against various attack variants and the adaptive attacks in Section~\ref{Sec:Robustness}.

In this section, we evaluate \name{} %compare \name{} with SentiNet~\cite{chou2018sentinet} 
%on the representative localized adversarial examples (i.e., \glassesattack{} and \patchattack{}) 
in two typical scenarios with multiple attack variants. 
%Our method can achieve the state-of-the-art results in both scenarios. %We also present that \name{} is robust in the physical world. To further demonstrate the robustness of \name{}, we extensively evaluate it against various attack variants and the adaptive attacks in Section~\ref{Sec:Robustness}.

% In Section \ref{Sec:Robustness}, we 
\subsection{Experiment Setup}
\label{Experiment Settings}
To evaluate the effectiveness of \name{},
%and show if it effectively achieves reproducible results, 
we use two typical scenarios, i.e., scene classification and face recognition. %, to demonstrate the effectiveness of \name{}. %we use two typical scenarios,  scenarios. 
In particular,
we construct universal and partial attacks by generating \patchattack{} in the scene classification scenario, and construct partial attacks by generating \glassesattack{} in the face recognition scenario. 

\noindent \textbf{Scene Classification.} ImageNet dataset \cite{deng2009imagenet} is commonly-used in large-scale scene classification tasks~\cite{krizhevsky2012imagenet}. 
It contains over 14 million images belonging to 1000 labels. Currently, there are a number of public models for scene classification that perform well on ImageNet dataset. Here, we use a pre-trained VGG16 model~\cite{simonyan2014very} that achieves over 92\% top-5 test accuracy in ImageNet. To evaluate TaintRadar, we implement localized universal  attacks~\cite{brown2017adversarial} and develop partial attacks using the same methodology. We also conduct the corresponding physical attacks. 
%follow the methodology of the universal localized attack method in~\cite{brown2017adversarial} and }
%we follow the localized attack method in~\cite{brown2017adversarial} 
%and construct adversarial patches to obtain enough adversarial images.

\noindent \textbf{Face Recognition.}
To evaluate TaintRadar in the face recognition scenario in the digital world, we follow the experiment settings  in~\cite{sharif2016accessorize} and use VGGFace dataset~\cite{parkhi2015deep}. The VGGFace dataset is a large scale dataset, consisting of 2.6 million images of  2,622 celebrities. Besides, we use the pre-trained VGG-Face model proposed in~\cite{parkhi2015deep}, achieving over 98\% test accuracy 
on the VGGFace dataset. We generate adversarial glasses to assess the performance of our approach. Moreover, to effectively evaluate TaintRadar in the real world, we retrain a new model that can recognize three authors of this paper and 140 celebrities in the PubFig dataset~\cite{kumar2009attribute} using transfer learning~\cite{yosinski2014transferable}.
%and the images are all collected from the internet. 
% Similar to~\cite{sharif2016accessorize}, we use transfer learning~\cite{yosinski2014transferable} in the attacks, which preserves the convolutional layers in CNN model~\cite{parkhi2015deep} and updates the fully connected layers only.
%we follow the settings in~\cite{sharif2016accessorize} to  identify the first three authors of this paper and 140 celebrities in the PubFig evaluation dataset~\cite{kumar2009attribute}. 
%PubFig is a large, real-world face dataset and the images are all collected from the internet. We retrain a new model to recognize these 143 identifies based on the idea of transfer learning~\cite{yosinski2014transferable}. 
%We preserve the convolutional layers in CNN model~\cite{parkhi2015deep} for feature extraction, modify and retrain the fully connected layers only.

% The scenarios for localized adversarial attacks mainly include scene classification and face recognition. 
% %In our experiments, we use ImageNet~\cite{deng2009imagenet} and VGGFace~\cite{parkhi2015deep} respectively to evaluate TaintRadar. 
% The specific information about datasets and models are shown in Table~\ref{experiment settings}.

We %\ft{use experiments to -> first} 
first show how we set the parameters in \name{} and then evaluate \name{} by measuring the true positive rate (TPR) and the false positive rate (FPR) under various settings in both digital and physical worlds.

%In experiments, we leverage inert patterns shown in Figure~\ref{Examples of inert patterns} to serve as the filling content for each region removal in each dataset.

%\ftdel{We measure the attack success rate with and without TaintRadar.} To evaluate the effectiveness of TaintRadar, we collect the true positive rate (TPR) and the false positive rate (FPR) under each attack setting during the process. Moreover, the attack success rate with and without \name{} will be evaluated to show the robustness of a model with TaintRadar under various attacks.

\subsection{Selection of Parameters in \name{}}

\begin{figure}[ht]
\vspace{-0.10in}
	\centering
	\subfloat{
	\label{fig:changeOfFPR} 
		\begin{minipage}[t]{0.4\linewidth}
			\centering
	\includegraphics[width=1.2in]{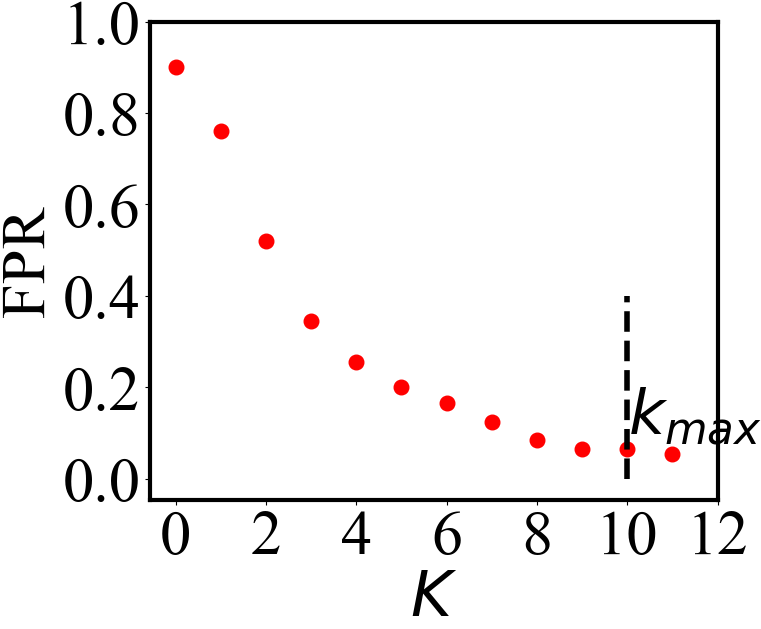}
			%\caption{fig1}
		\end{minipage}%
	}%
	\subfloat{
	\label{fig:relevance} 
		\begin{minipage}[t]{0.4\linewidth}
			\centering
	\includegraphics[width=1.17in]{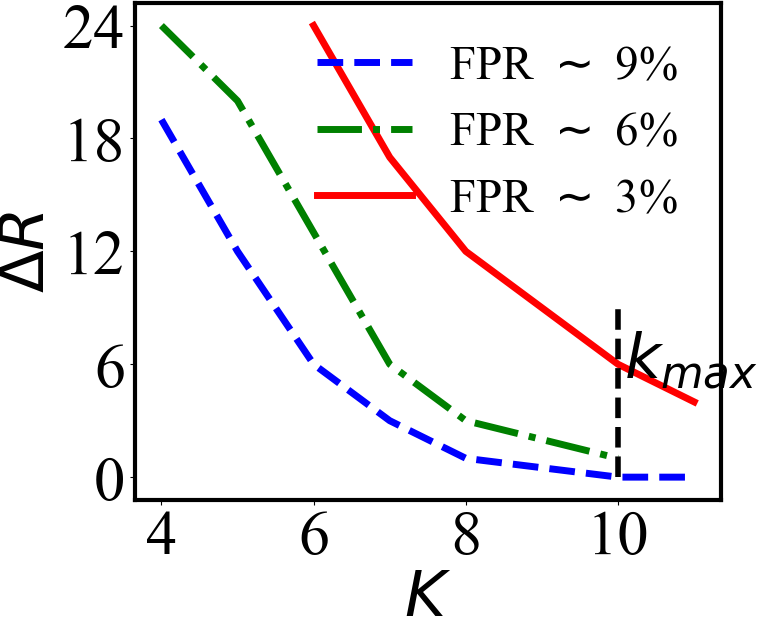}
			%\caption{fig2}
		\end{minipage}%
	}%
%	\vspace{-0.05in}
\\
\white{...} (a) 
\white{...............................} (b)
	\centering
	\vspace{-0.1in}
	\caption{(a) The changes of FPR as $K$ increases; (b) The correlation among $\Delta R$, $K$, and FPR.}
% 	\label{fig:RelevanceTPandFP}
    \vspace{-0.2in}
\end{figure}

%\name{} aims to detect the attack that is agnostic in advance, \ft{which means} it does not require any prior knowledge of the adversarial attack. Therefore, we consider to decide the parameters based on a set of benign images. 
To make \name{} attack-agnostic (i.e., requiring no prior knowledge of adversarial examples), we use a set of benign images to decide two parameters in \name{}, i.e., top-$K$ and $ \Delta {R}$. Since our experiments show that the settings in different scenarios are similar, for simplicity, we take scene classification as an example to illustrate the parameter selection. First, we randomly select 200 images from the ImageNet Validation Dataset and run \name{} with different $\Delta R$ and $K$ combinations. 
The results are shown in Figure~\ref{fig:changeOfFPR} and Figure~\ref{fig:relevance}. 
%empirical examples are shown in Figure . Since the result points are concrete in space, to better illustrate the trends, we fit a curve for each FPR interval in Figure 
%
%
%In particular, when $K$ is set too small, the intersection region found by \name{} may be too large for benign images, which can result in a high FPR. As $K$ increases, the FPR will decrease. However, the intersection region of adversarial images gets smaller gradually, and the attacked region may not be found completely. We choose $K$ and $\Delta R$ by the empirical study of a benign test dataset  (see Figure~\ref{fig:parameters})\ft{problem here: the figs shown in Figure 5 are ideal trends of the experimental result, and they do not depict any real result. This condition should be explained clearly.}
%
%
%\ftadd{
%We theoretically analyze how $K$ and $\Delta R$ influence detection performance (i.e., TPR and FPR).
%}
We find that when $K$ is set too small, the intersection region found by \name{} may be too large for benign images. If removing the region, the ranking changes of the original labels would be large which result in a high FPR. %\lxk{as shown in Figure~\ref{fig:changeOfFPR}}.
%Accordingly, for a fixed $\Delta R$, FPR would be high.
As shown in Figure~\ref{fig:changeOfFPR}, when $K$ increases, the FPR will decrease. However, the intersection region of adversarial images gets smaller gradually, and the attacked region can not be found effectively as $K$ increases to a certain level, leading to a low TPR. Thus, we set threshold $k_{max}$ as the maximum value of $K$ to trade off between FPR and TPR, where FPR is close to the minimum value and meanwhile the maximum value $K$ ensures a highr TPR (see Section~\ref{subsec:detection}). 
%, where FPR stops declining and larger  might lead to lower TPR.

Moreover, we analyze the ranking changes of the original label after removing the intersection region so that we can choose $\Delta R$. In particular, when the ranking changes are larger than the threshold $\Delta R$, we can safely detect them as adversarial images, which means TPR would be high if $\Delta R$ is small.  
For benign images, the correlations among $\Delta R$, $K$ and FPR is shown in Figure~\ref{fig:relevance}. We can see that when FPR is given, the corresponding $K$ decreases as $\Delta R$ increases. Therefore, we can select an acceptable FPR for benign images in advance, e.g., 6\%. At the same time, $\Delta R$ and $K$ are mutually constrained to maintain a FPR value for benign images. Considering that the change of $K$ within a certain range has a small effect on the detection capability against adversarial images, we select $\Delta R$ as small as possible to improve the detection accuracy of \name{}.

\subsection{Digital-World Experimental Results}
\label{DigitalAttacks}
We evaluate the effectiveness of \name{} in the digital-world where an attacker can manipulate each victim image on the per-pixel level. We systematically measure the difficulty of generating an attack image under our defense. It is relatively easier for an attacker to deceive the classifiers without physical constraints. An attacker who cannot succeed in the digital world will mostly fail in the physical world. 

\begin{figure}[t]
\centering
%\vspace{-0.10in}
\begin{minipage}[t]{0.13\linewidth}
\centering
\includegraphics[width=0.45in]{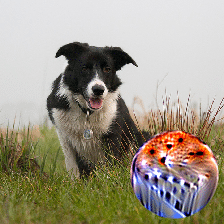}
%\caption{fig1}
\end{minipage}%
\begin{minipage}[t]{0.13\linewidth}
\centering
\includegraphics[width=0.45in]{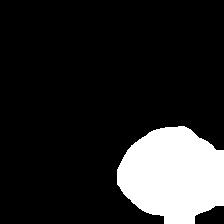}
%\caption{fig2}
\end{minipage}%
\begin{minipage}[t]{0.13\linewidth}
\centering
\includegraphics[width=0.45in]{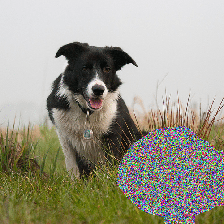}
%\caption{fig1}
\end{minipage}%
\begin{minipage}[t]{0.13\linewidth}
\centering
\includegraphics[width=0.45in]{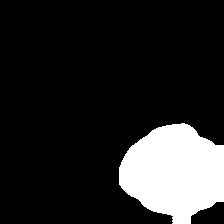}
%\caption{fig1}
\end{minipage}%
\begin{minipage}[t]{0.13\linewidth}
\centering
\includegraphics[width=0.45in]{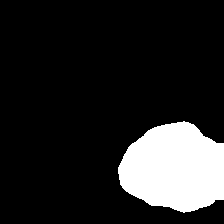}
%\caption{fig2}
\end{minipage}%
\begin{minipage}[t]{0.13\linewidth}
\centering
\includegraphics[width=0.45in]{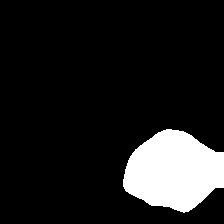}
%\caption{fig2}
\end{minipage}%
\begin{minipage}[t]{0.13\linewidth}
\centering
\includegraphics[width=0.45in]{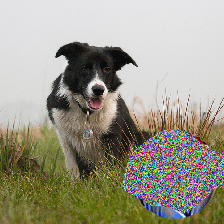}
%\caption{fig2}
\end{minipage}%

\begin{minipage}[t]{0.13\linewidth}
\centering
\includegraphics[width=0.45in]{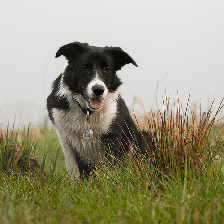}
%\caption{fig1}
\end{minipage}%
\begin{minipage}[t]{0.13\linewidth}
\centering
\includegraphics[width=0.45in]{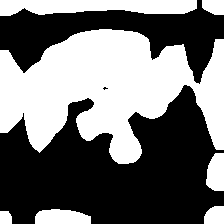}
%\caption{fig2}
\end{minipage}%
\begin{minipage}[t]{0.13\linewidth}
\centering
\includegraphics[width=0.45in]{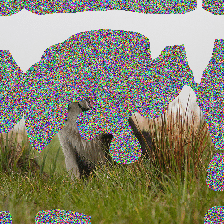}
%\caption{fig1}
\end{minipage}%
\begin{minipage}[t]{0.13\linewidth}
\centering
\includegraphics[width=0.45in]{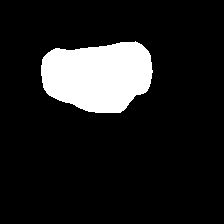}
%\caption{fig1}
\end{minipage}%
\begin{minipage}[t]{0.13\linewidth}
\centering
\includegraphics[width=0.45in]{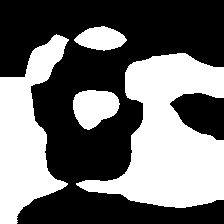}
%\caption{fig2}
\end{minipage}%
\begin{minipage}[t]{0.13\linewidth}
\centering
\includegraphics[width=0.45in]{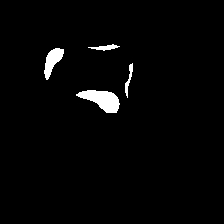}
%\caption{fig2}
\end{minipage}%
\begin{minipage}[t]{0.13\linewidth}
\centering
\includegraphics[width=0.45in]{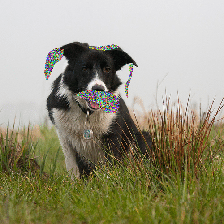}
%\caption{fig2}
\end{minipage}%
\centering
\vspace{-0.1in}
\caption{The detection results of each  TaintRadar step. Column 1 shows the benign and the victim input images, Columns 2 and 3 show the estimated critical regions and the intermediate inputs filled with filling patterns, Columns 4 and 5 illustrate two examples of sub-masks out of four,  
Columns 6 and 7 show the intersection regions of these sub-masks and final inputs.}
\label{fig:Regionproposal}
\vspace{-0.2in}
\end{figure}

\noindent \textbf{Scene Classification.}
\label{exp:scene}
%Adversarial Patches}
%\paragraph{Attack Variables}
To show that our defense is effective and robust, we construct both partial and universal attacks with varying attack settings when generating patches. The detailed attack has three settings. First, the batch size in patch generation is the number of images used to generate patches, showing the generalization capability of patches. For example, a patch created on 16 images has a more universal effect on held-out images than a generated patch for only 1 image. Second, for one size or multiple sizes in patch generation, an adversarial patch generated with multiple sizes is more robust against different shooting distance than those generated with one size. Third, we change the size of patches in victim images to estimate the effectiveness of \name{} against patches of different sizes. Lastly, the patch position in attacking is another factor to evaluate if the information loss of the original object will influence the detection performance of \name{}.

In this experiment, we randomly select 400 images from the ImageNet Validation Dataset as the benign image dataset. These 400 images are all correctly classified as ground truth labels. % because both an attack and FPR evaluated on wrongly predicted images are meaningless. 
With the benign images dataset, similar to~\cite{brown2017adversarial}, we perform targeted attacks by using the cleverhans code~\cite{papernot2018cleverhans} to construct patches. The attacks make the victims misclassified as a \emph{toaster} (the corresponding label is 859). According to the method of setting parameters above, we use a randomly selected held-out benign dataset containing 200 images from validation set. We set the expected FPR as 6\%, and the generated parameters are 9 for $K$ and 2 for $\Delta R$. The FPR evaluated on 400 held-out benign images is 6.25\%.

\begin{table*}[t]
\caption{The first column of each setting is the successfully-generated rate (SR) out of 400 attempts. The other two show the TPRs of SentiNet and \name{} against adversarial patches. All numbers are shown in their percentages.}
\label{table:resultspatch}
\vspace{-0.09in}
\resizebox{\textwidth}{!}{ 
\begin{tabular}{cc|ccc|ccc|ccc|ccc|ccc|ccc}
\toprule[1.5pt]
 &  & \multicolumn{9}{c|}{\textbf{Single Size}} & \multicolumn{9}{c}{\textbf{Multiple Sizes}} \\ \midrule[1.5pt]
 & \multicolumn{1}{|c|}{\textbf{Batch Size}} & \multicolumn{3}{c|}{1} & \multicolumn{3}{c|}{4} & \multicolumn{3}{c|}{16} & \multicolumn{3}{c|}{1} & \multicolumn{3}{c|}{4} & \multicolumn{3}{c}{16} \\
 \cline{2-20}
\textbf{Position} & \multicolumn{1}{|c|}{\textbf{Patch Size}} &SR & SentiNet & \textbf{Ours} &SR & SentiNet & \textbf{Ours} &SR & SentiNet  & \multicolumn{1}{c|}{\textbf{Ours}}&SR & SentiNet & \textbf{Ours}&SR & SentiNet & \textbf{Ours}&SR & SentiNet & \textbf{Ours}   \\ \hline
\multicolumn{1}{c|}{\multirow{3}{*}{\textbf{\begin{tabular}[c]{@{}c@{}}Right-\\ bottom\end{tabular}}}} & 0.2 &37.75 &  50.33 & \textbf{97.96} &29.00 &  78.38  & \textbf{99.08} &27.75 & 87.93&  \multicolumn{1}{c|}{\textbf{98.25}}  &6.50 &  3.85 &\textbf{80.00} &6.00 &  21.05  & \textbf{100.00} &4.75 & 12.50 & \textbf{95.83} \\

\multicolumn{1}{c|}{} & 0.3 &97.50 &  81.28 &\textbf{97.95} &97.25 &  99.48 & \textbf{99.74} &96.00 &\textbf{100.00} & \multicolumn{1}{c|}{97.94} &78.75 &  58.41 & \textbf{95.83} &83.25 &  98.76 & \textbf{99.07} &80.75 & \textbf{99.40}& 97.29 \\

\multicolumn{1}{c|}{} & 0.4 &100.00  &  57.25 &\textbf{91.50} &100.00 &  \textbf{97.49} &97.24 &99.50 & \textbf{99.75} & \multicolumn{1}{c|}{97.25} &96.00 &  46.88 &\textbf{86.98} &98.00 &  \textbf{94.39} &94.13 &9800 & \textbf{98.98} & 88.27 \\ %\cline{1-14} 
\hline
\multicolumn{1}{c|}{\multirow{3}{*}{\textbf{Random}}} & 0.2 &58.50 &  29.91 &\textbf{95.26} &38.50 &  48.21 &\textbf{96.39} &42.00  & 56.49 & \multicolumn{1}{c|}{\textbf{95.45}} &15.50  &  29.91 &\textbf{79.03} &12.50 &  48.21 &\textbf{91.30} &11.75 &56.49 & \textbf{81.63} \\
\multicolumn{1}{c|}{} & 0.3 &98.75 &  58.99 &\textbf{95.19} &97.00  &  87.47 &\textbf{98.47} &97.75  & 88.92 & \multicolumn{1}{c|}{\textbf{95.10}} &86.00  &  58.99 &\textbf{93.88} &88.00 &  87.47 &\textbf{96.36} &89.75 & 88.92 & \textbf{95.14} \\
\multicolumn{1}{c|}{} & 0.4 &100.00 &  34.75 &\textbf{88.25} &99.75 &  66.25 &\textbf{95.75} &100.00  & 8.25 & \multicolumn{1}{c|}{\textbf{71.18}} &97.75  &  34.75 &\textbf{82.10} &98.75  &  66.25 &\textbf{88.92} &99.25 & 71.18 &  \textbf{80.76}\\ 
\bottomrule[1.5pt]
\end{tabular}
}
\vspace{-0.15in}
\end{table*}

A pair of examples is shown in Figure~\ref{fig:Regionproposal}. The negative regions focus on the same area in the flow of malicious image detection, while for the benign example, they are scattered. From Table~\ref{table:resultspatch}, we can see that TaintRadar achieves a high TPR under different settings, which ensures the final success rate of patch attacks at a low level. %Despite this, 
The TPR decreases slightly when patches are placed at a random position in victim images. The reason is that they are likely to cover the main object in the original image, which incurs information loss of the original object and leads to a performance degradation of the detection. 
%However, TaintRadar still achieves a high TPR under this setting. 

%As shown in Table~\ref{table:resultspatch}, TaintRadar maintains a high detection rate under different settings, which ensures the final success rate (FSR) of patch attacks at a low level. Despite this, the detection rate decreases slightly when patches are placed at a random position in the image. The main reason is that they are likely to cover the main object in the original image, which incurs information loss of the original object and leads to a performance degradation of the detection. However, TaintRadar still achieves a high detection rate under this setting. 

\begin{figure}[ht]
\vspace{-0.20in}
	\centering
	\subfloat[Adversarial patches]{
	\label{fig:TPR_patch}
		\begin{minipage}[t]{0.4\linewidth}
			\centering
	\includegraphics[width=1.04in]{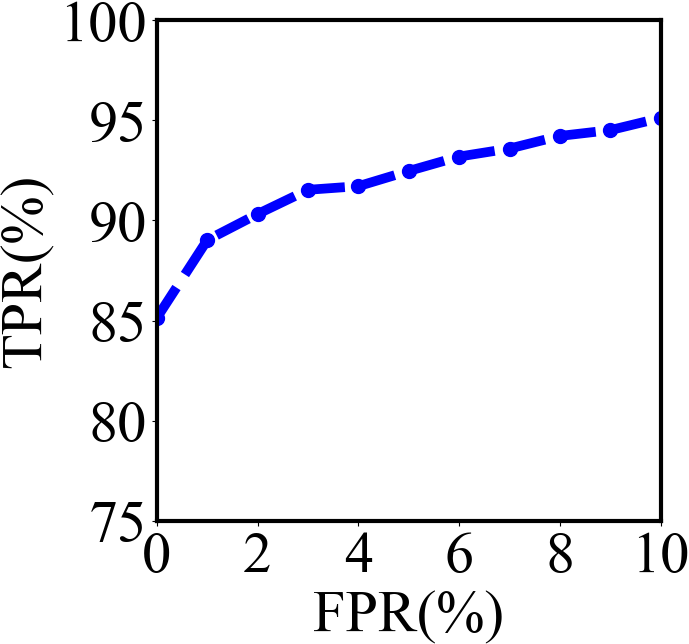}
			%\caption{fig1}
		\end{minipage}%
	}%
	\subfloat[Eye-glasses attacks]{
	\label{fig:TPR_glasses}
		\begin{minipage}[t]{0.4\linewidth}
			\centering
	\includegraphics[width=1in]{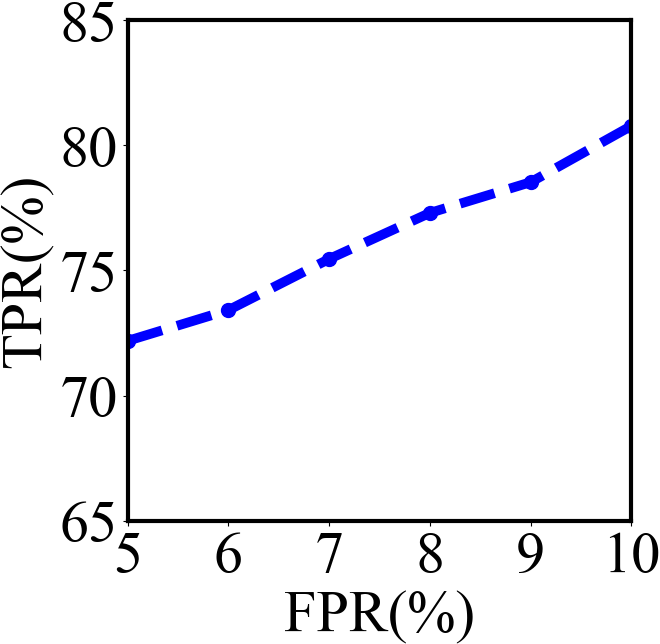}
			%\caption{fig2}
		\end{minipage}%
	}%
	\vspace{-0.10in}
	\centering
	\caption{TPR of \name{} under different FPR settings. }
	\label{fig:RelevanceTPandFP}
\end{figure}

Meanwhile, our approach is robust to different sizes of patches. There are a few abnormal results when the patch is generated with multiple sizes and the patch size in victim images is 0.2. The reason is that the number of adversarial images that are attacked successfully is really small. Therefore, there may exist deviations in the detection results with few samples.
%Note that, TaintRadar is a generic defense system that can be applied against various attacks with different generalization capabilities. 
Also, we find that \name{} can detect both partial (batch size is 1) and universal attacks.
%However, existing defenses against localized adversarial attacks are invalid for partial attacks, e.g., the batch size is 1. Moreover, 
In addition, the performances are stable in different settings when a smaller FPR is used. For example, when the FPR changes from 6\% to 3\%, the drops of performance range between 0\% and 4\% in Table~\ref{table:resultspatch}. Figure ~\ref{fig:TPR_patch} shows the averaged trend under different FPR settings.

We reproduce SentiNet following the settings in~\cite{chou2018sentinet}. It shows a similar 4\% FPR and a 98.11\% TPR on the adversarial patch task in~\cite{chou2018sentinet}. As shown in Table~\ref{table:resultspatch}, using the same estimated curve and benign test set, SentiNet also achieves comparable results when the \patch{} have strong generalization capabilities. However, when the batch size used in generation drops, the performances depicts a clear trend of degradation. Also, we note that the performance peaks at patch size 0.3. The reason behind is that patch of 0.2 has less chance of fooling the majority of benign sets, which is consistent with the successfully-generated rate shown in Table~\ref{table:resultspatch}. For patches with a size of 0.4, the average confidence drops compared to the size of 0.3, leading the output points cross the decision boundary. In total, \name{} achieves a better and more stable performance in different attack settings, including generalization capability, patch size, and patch position.
% Due to the page limit, we show the results with a lower FPR (i.e. 3\%) in  Appendix~\ref{appendix:deltailPerformance}.

\vspace{-0.15in}
% \begin{figure}[ht]
% %\vspace{-0.10in}
% \centering
% \subfloat{
% \begin{minipage}[t]{0.18\linewidth}
% \centering
% \includegraphics[width=0.65in]{figs/ban_att1.jpg}
% %\caption{fig1}
% \end{minipage}%
% }%
% \subfloat{
% \begin{minipage}[t]{0.18\linewidth}
% \centering
% \includegraphics[width=0.65in]{figs/ban_att2.jpg}
% %\caption{fig2}
% \end{minipage}%
% }%
% \subfloat{
% \begin{minipage}[t]{0.18\linewidth}
% \centering
% \includegraphics[width=0.65in]{figs/ban_att3.jpg}
% %\caption{fig1}
% \end{minipage}%
% }%
% \subfloat{
% \begin{minipage}[t]{0.18\linewidth}
% \centering
% \includegraphics[width=0.65in]{figs/ban_att4.jpg}
% %\caption{fig2}
% \end{minipage}%
% }%
% \subfloat{
% \begin{minipage}[t]{0.18\linewidth}
% \centering
% \includegraphics[width=0.65in]{figs/ban_att5.jpg}
% %\caption{fig2}
% \end{minipage}%
% }%
% %\vspace{-0.15in}
% \centering
% \caption{The patch attacks in the real world. We place the printed patch at different positions and angles to make the banana misclassified, then use video frames to evaluate the performance of our approach.}
% \label{fig:physicalBanana}
% \end{figure}

\noindent \textbf{Face Recognition.}
%\paragraph{Attack Variables}
In this scenario, we construct eye-glasses attacks, i.e., generating adversarial eys-glasses.  All of them are partial attacks, since each pair of glasses target one person.
%We do not have variables the same to that in the scene classification scenario. Note that, glasses are of different shapes with ~\patchattack{}, we also test the robustness of \name{} on different shapes. For simplicity, similar to~\cite{sharif2016accessorize}, we use the partial attack to demonstrate it performance.
%
%\paragraph{Datasets}
To generate adversarial eye-glasses, we set victim image dataset  by randomly sampling 20 from 2622 people and each with 50 aligned images (totally 1000 images).
%, and the same to the experiments above. 
%For each target, we randomly select 50 aligned images that are correctly classified by the VGG-Face model, totally 1000 images. 
Then, we launch the attack targeting at \emph{A.J Buckley} (the corresponding label is 0) with these images. We set the stop probability of 0.95 and with maximum 300 iterations following the settings in~\cite{sharif2016accessorize}. 
To determine parameters, we randomly select a held-out benign dataset with 20 people each with 10 images. Also, with the FPR set to 6\%, we get parameters set as 13 and 13, for $K$ and $\Delta R$, respectively. The filling pattern here is the average face 
% (see Figure~\ref{aveface}) 
generated with all the VGGFace data, which preserves the facial information.

%\paragraph{Results}
% The last two rows in Figure~\ref{fig:Regionproposal} illustrate the different behavior of \name{} on malicious and benign images. 
With 978 successfully attacked images out of 1000 attempted ones, the TPR of \name{} is 73.42\%. The FPR evaluated on 200 held-out benign images is 5.50\%. The TPR is mainly affected by two factors. First, the region found by \name{} is accumulated on the face area, which is relatively large for benign images. Second, the region covered by eye-glasses  is critical for the VGG-Face model; however, the overlapping part of the original semantics will only impair the confidence of the original object’s related labels. As our experiment shows, even when a large part of the critical regions is covered, it still has a minor impact on the final result. It conforms to the detection results of randomly placed patches as discussed in Section~\ref{exp:scene}, e.g., covering 20\% of the critical region can still yield a 94\% TPR. 
%Since we use face dataset, the region found by \name{} accumulated on the face area and is relatively large for benign images. 
%Besides, similar to randomly placed patches discussed in Section~\ref{exp:scene}, the region of eye-glasses covers is critical for the VGG-Face model, and thus it can damage the face information in the original image. Thus, the detection effectiveness of TaintRadar is  constrained. 
We discuss how to improve the detection performance in Section~\ref{GAN-Based-inpaint}. Also,  SentiNet~\cite{chou2018sentinet} achieves a 2\% FPR, while only 0.72\% of these 978 images are flagged as malicious with a nearly 0 fooled rate on the benign test set. Again, it proves our method generalizes better than SentiNet on localized partial attacks.

\vspace{-0.05in}
\subsection{Physical-World Experimental Results}
\label{PhysicalAttacks}
% The localized adversarial examples such as adversarial patch and \glassesattack{} can be printed out to raise a tremendous physical-world impact, which are more harmful than the attacks generating imperceptible adversarial perturbations. 

%We show that \name{} is also effective in the real world.

%, we reproduce these two attacks in the real-world and validate the robustness of \name{}. 
%In this experiment, we introduce the physical experiment setup of two kinds of attacks respectively, and assess the performance of \name{} in the physical environment.
%
% We collect images of subjects using a Olympus Tough TG-5 camera. After patch generation, all \patch{} are printed with HP Color Laser Jet Pro M154nw.  \name{} is evaluated by using videos. To achieve comparable attack results, all videos are collected in a room without exterior windows and with a  stable lighting condition.

%\subsubsection{Adversarial Patch}
\noindent \textbf{Scene Classification. }
%For the scene classification scenaior in the physical environment, 
We use the pre-trained VGG16 model in the physical world  same as that in the digital world. Following the setting of~\cite{brown2017adversarial}, we captured 5 videos of a banana with and without a printed patch next to it, respectively. 
The relative position and the orientation of the banana and patch varies between each video. 
The VGG16 model without using \name{} classifies all 600 benign video frames (120 frames each) correctly and 598 out of 600 video frames with the printed patch as the target label \emph{toaster}. By using the same parameters in the digital world, \name{} achieves 88.63\% TPR on 598 victim image and 0.17\% FPR on 600 benign inputs.
%Also, the banana is correctly classified in 99\%  without the presence of the attack. 
% The result demonstrates that \name{} effectively detect physical-world attacks in the scene classification scenario. 
In real deployment, the TPR can be further improved by verifying the prediction consistency of the contiguous frames, for example, using 5 contiguous frames to judge if a scene is under an attack during a short period of time.

%\subsubsection{Eye-Glasses Attack}
\noindent \textbf{Face Recognition. } 
To construct partial attacks in the face recognition scenario, i.e., eye-glasses attack, 
% three authors of this paper impersonate different targets, respectively. \ft{As shown in Figure~\ref{fig:physicalGlasses}, we generate malicious eye-glasses for each person.} \ft{The experiment setting is depicted in Table~\ref{table:physicalglasses}}. 
we use the similar setting in the training of $DNN_{C}$ in~\cite{sharif2016accessorize}, which identifies 143 subjects composed by the first three authors of this paper and 140 celebrities from the PubFig evaluation dataset, each person with 40 training images. Based on the idea of transfer learning~\cite{yosinski2014transferable}, we keep the first 37 layers of the VGG-Face Model~\cite{parkhi2015deep}, and append a fully connected layer with 143 output neurons followed by a softmax layer\footnote{~\cite{sharif2016accessorize} appended an extra sigmoid layer before the softmax. However, we think this was a typo. After normalized by a sigmoid, the final confidence of each label could not be greater than $\frac{e^{\prime}}{\sum_{i \neq t} e^{0}+e^{1}} \approx \frac{2.718}{142+2.718} \approx 0.0188$.}, obtaining a 94.93\% accuracy on a held-out test set. Aiming at this model, we train a pair of eye-glasses against 30 benign images for each of the three authors. Videos containing 386, 309 and 410 
frames for each author are collected, with 4.94\%, 93.85\% and 84.54\% successful rates. Among these successful attacks, 68.42\%, 79.31\% and 80.83\% are successfully detected by \name{}, with a 79.78\% TPR in total. We achieve a better result than on the VGG-Face Model. The reason is that the model trained on a small dataset has less impact on benign images. However, the estimated critical region and the intersection region of the VGG-Face Model~\cite{parkhi2015deep} are relatively large, leading to a higher FPR. This experiment result demonstrates that \name{} is also robust on the eye-glasses attack in the real world.

\subsection{Run-time Computation Overhead}
\label{RuntimeOverhead}
For deployment in a real-time neural network system, e.g., surveillance, the run-time overhead of \name{} should be small. Our experiments are executed on an RTX 2080Ti GPU with an i7-6850K CPU. We average the run-time computation delay over processing 100 images on VGG16 for ImageNet, which shares the same structure with VGG-Face.
%, to show the efficiency of \name{} and the impact of parameter $K$. The result is shown in 
% Figure~\ref{fig:runtime} shows the efficiency of \name{} under the impact of parameter $K$.
%
% We can see that \name{} incurs an extra delay of around 4ms for each increase of $K$. 
The overhead is about 79ms for each image when the $K$ value set to 9. For each increase of $K$, \name{} incurs an extra delay of around 4ms. Actually, the delay can be further reduced by migrating the CPU computations to GPU or performing parallel computations. This result is around 100X faster than SentiNet~\cite{chou2018sentinet} that requires around 7.73s for each image under the same environment. Therefore, \name{} is readily deployable in real time-bounded DNN systems.
% \begin{figure}[h] 
% 	\centering 
% %	\vspace{-0.10in}
% 	\includegraphics[width=1.7in]{./figs/time.png}
% 	%\vspace{-0.05in}
% 	\caption{Run-time overhead.} 
% 	\label{fig:runtime} 
% \end{figure}
%!TEX root=./main.tex
\vspace{-0.08in}
\section{Robustness Evaluation} %Analysis}
\label{Sec:Robustness}
We evaluate the robustness of \name{} against two types of advanced attacks. First, the attackers use patches of different numbers or shapes to construct adversarial variants. Second, the attackers conduct adaptive white-box attacks using  
prior knowledge of \name{}. We study the attacks on scene classification tasks with the VGG16 model, including both universal and localized attacks. The effectiveness of \name{} against these attacks can be generalized to attacks over other tasks. In the experiments, we set FPR as 6\% and use random noise as filling pattern the same as in Section \ref{sec:Experiments}.

\subsection{Robustness against Adversarial Variants}
\label{Sec:robustnessVariants}
%\subsubsection{Multiple patches}

%We evaluate two types of adversarial variants: {\em multiple patches} and {\em  single patch of different shapes}.

\noindent \textbf{Attacks Using Multiple Patches. }
%The attacks discussed above generate contiguous regions with only one adversarial patch in an image. 
To validate if \name{} can identify more than one critical region, we launch attacks using multiple patches. We construct 25 patches that target at mislabelling 400 benign images as \emph{toaster}, i.e., one patch for 16 images.
%construct two patches on the left and the right bottom of the victim images, respectively. targeting at the class toaster 
When the same patch, with a size of 0.4, is applied twice on the left and the right corners of victim images, all victim images are successfully misclassified as the target label. % Figure~\ref{fig:attackVariants} shows an example of victim images.
%All of the generated images succeed to fool the model. 
%An Example is shown in Figure~\ref{Multiple patches attack} on the left. 
Our experiments show that \name{} can achieve 91.5\% TPR on detecting adversarial examples with multiple patches.

%\subsubsection{Different shapes}
\noindent \textbf{Patches of Different Shapes. }
We also evaluate the effectiveness of \name{} on detecting adversarial patches of different shapes. We construct star-shaped and lightning-shaped patches as examples. % (see Figure~\ref{fig:attackVariants}). 
%to evaluate TaintRadar. 
We randomly select 200 images from  ImageNet and fool the VGG16 model to misclassify them as \emph{toaster}. 
To avoid covering main objects of original images, we put the patch on the bottom right of victim images. 
We successfully generate 156 adversarial images each with a star-shaped advesarial object and 158 adversarial images each with a lightning-shaped adversarial object.
%Examples are depicted in Figure~\ref{fig:differentshape}, where the victim images on the left and the final mask highlighted in green on the right.
Among them, TaintRadar achieves 92.31\% and 90.5\% TPRs for star-shaped and lightning-shaped adversarial objects, respectively. 

\begin{figure}[t]
	\centering 
	\vspace{-0.05in}
	%\vspace{1cm}
	\includegraphics[width=0.59\linewidth]{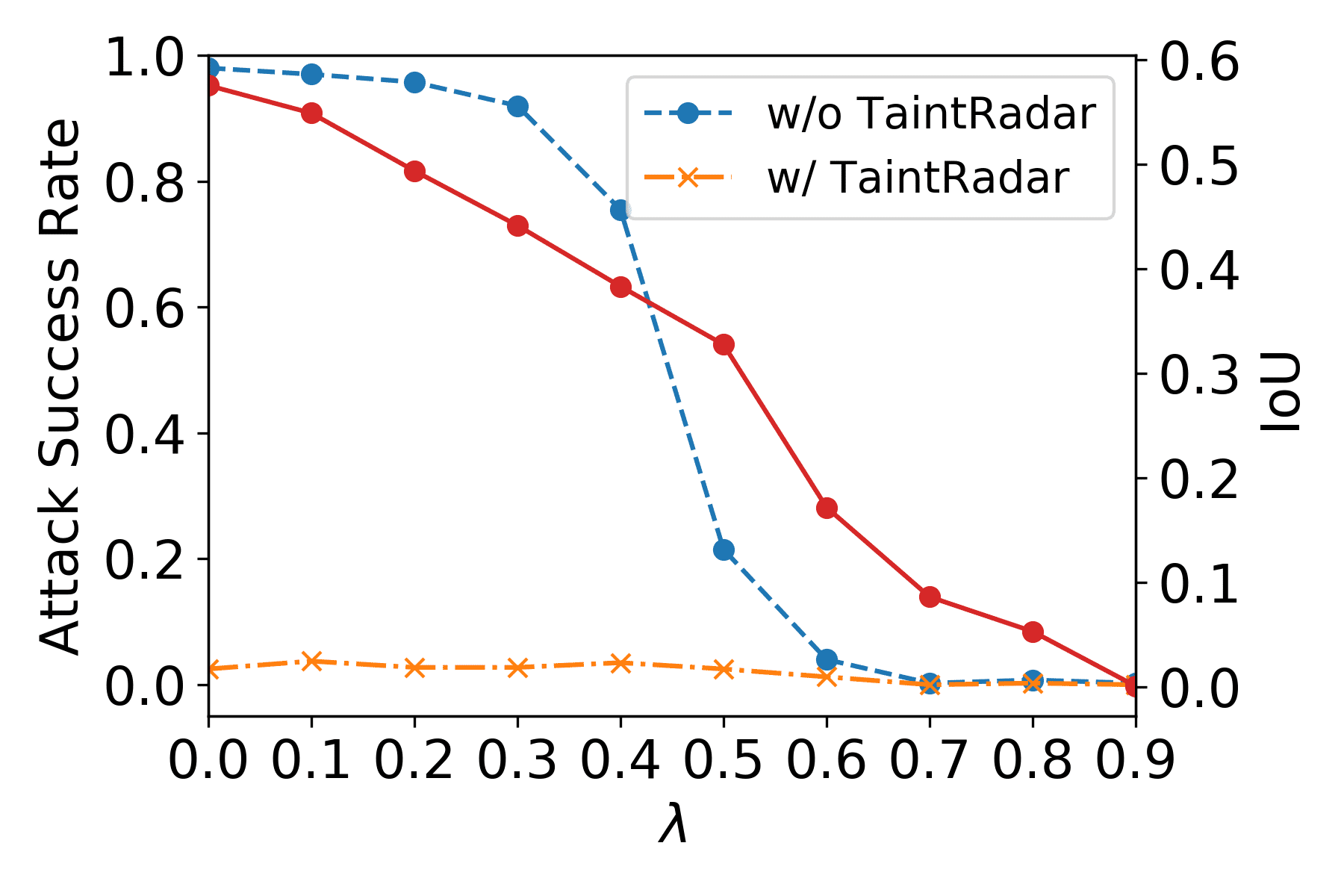}
	\vspace{-0.20in}
	\caption{The impact of region misleading on \name{}. %As $\lambda$ grows, the optimization focuses more on misleading the critical region estimation. 
	Solid line shows the averaged IoU value of generated attacks; dashed lines show the attack success rates with or without \name{}.}
	\label{fig:regionMislead}
	\vspace{-0.22in}
\end{figure}
\vspace{-0.08in}

\subsection{Robustness against Adaptive Attacks}
\label{Sec:adaptive attacks}
% The experimental results demonstrate the \ft{genericity} of ~\name{}.
We show the robustness of \name{} against adaptive attacks that have white-box knowledge of TaintRadar. 
Specifically, %they attack the two major steps of our method, i.e., 
an attacker can choose to mislead critical region estimation of \name{} or manipulate the rankings of target labels for fooling the critical region based detection of \name{}. Note that, partial attacks can be seen as an adaptive attack against defense methods that only consider universal attacks.
%(which are used in the detection step). 
%, where the first estimates the critical region in the image, and the second detects whether the critical region is malicious by critical region analysis. To comprehensively evaluate the robustness of our approach, we consider adaptive attacks against these two steps in ~\name{}, i.e., critical region estimation and \ft{malicious region detection}, respectively. 

%\subsubsection{Attacking the region estimation}
\noindent \textbf{Misleading Critical Region Estimation.}
%\name{} requires to estimate critical regions for each image, which may be manipulated by an adaptive attacker. 
%An adaptive attacker may attempt to mislead the estimation of critical region so that \name{} would identify the adversarial region inaccurately or even a completely different area. 
%
Inspired by a misleading attack that constructs imperceptible pixel-level perturbations~\cite{zhang2018interpretable}, 
%Here, 
%To validate the influence of such attacks on our method, 
we construct localized attacks that can interfere with \name{} by adding pixel-level perturbations in a limited area\footnote{The original attack proposed in~\cite{zhang2018interpretable} needs to change pixels in the entire images, which cannot be directly applied in localized attacks.}. 
The attack aims at deceiving CNN models and misleading estimated regions. % at the same time. 
Through our extensive experiments, we find that the attack does not have an impact on \name{} since it is hard for the attack to achieve both misclassification and region misleading. We use $1-\lambda$ and $\lambda$ to balance the optimization goal of misclassification and region misleading, respectively, where a larger $\lambda$ means the optimization focuses more on misleading the critical region estimation. 
% The formalized loss function and an example of attacked image  can be found in Appendix~\ref{appendix:regionDetails}. 
% Since , we are here to minimize over
% $\left(1-\lambda\right)\ell_{\mathrm{prd}}\left(f(x), c_{t}\right)-\lambda \ell_{\mathrm{est}}\left(e(x ; f), g_{t}\right)$, where $\ell_{\mathrm{prd}}$ is the prediction loss between the current state $f(x)$ and target $c_{t}$, and $\ell_{\mathrm{est}}$ is the loss that stands for the distance between the estimated area $e(x ; f)$ and the ground truth attacked region $g_{t}$. 
We generate 25 patches with varid $\lambda$ and apply them to 400 images for each $\lambda$. To show the accuracy of our critical region estimation under the misleading attacks, we use Intersection over Union (IoU), i.e.,  
$\frac{|E \cap G|}{|E \cup G|}$, where $E$ is the \name{}-estimated region and $G$ is the actual adversarial region.
Figure~\ref{fig:regionMislead} shows that the value of IoU drops as $\lambda$ increases, 
%which means the effect that the attack misleads our estimated area away from the adversarial region.
which means 
%the effect that 
the attack misleads our estimated area away from the actually adversarial region.
%However, the attack success rate without ~\name{} decreases at the same time, because the optimization process focuses less on \zxl{fooling the classification task.} 
At the same time, the attack success rate (without~\name{}) decreases, because the optimization process focuses less on fooling the classification task.
%misleading the region estimation. 
In other words, we find these two optimization tasks are contradictory to each other. When using \name{}, the attack success rate remains below 10\% all the time, illustrating that \name{} can effectively inhibit the adaptive attacks.
%The attack success rate with our approach remains below 10\% all the time. 
Moreover, we build two other adaptive attack methods which aim to minimize \name{}-estimated regions and make \name{} identify a pre-defined area as the critical region, respectively. \name{} shows similar effectiveness against those attacks. 

\noindent \textbf{Deceiving Critical Region Based Detection.}
%Robustness against Attacks on \ft{Malicious Region Detection}. } %Detection Mechanism}
%An attacker can 
%After the process of critical region estimation and analysis, we use a threshold $\Delta R$ to finally determine whether test images are malicious. With the full knowledge of our design, 
%
Attackers may attempt to evade the detection of \name{} by adaptively controlling the ranking changes of the target label to be lower than the threshold $\Delta R$. More specifically, attackers can select a target label with the highest confidence in victim images. We implement both universal and partial attacks as follows.
%An attacker with the full knowledge of the \name{} details can construct adaptive attacks to evade malicious region detection by making the ranking of test images lower than threshold $\Delta R$.  
%More specifically, an attacker might launch an attack by selecting a target label with the ranking as high as possible in the original image. 
%
%For instance, 

To construct the universal attacks, we collect images of different source-labels, average the confidence of these images, and then choose the label with the highest ranking (instead of using the original label) as the target label. Here, we generate 50 patches and apply each to 16 randomly sampled images. Among all 800 generated images, 766 images can successfully fool the VGG16 model and 91.78\% of them can be detected by \name{}. For the partial attacks, we collect images with the same source-label under different settings (e.g., light condition and background)  and set the target label as the one with the highest ranking, achieving 789 successful attack attempts. \name{} still achieves a 68.44\% detection rate. We see that the attacks do not have a significant impact on \name{} though some attacks can evade detection. The reason is that, in the presence of \name{}, the ability of the attacker is significantly weakened. First, the attack interfaces are greatly reduced under \name{} since an attacker can use a very limited number of labels to construct the attacks. For instance, an attacker can only use one label out of 1000 labels to construct the attack in our experiments. Second, in real world, it is difficult for  attackers to accurately determine the target label since the rankings of the target label are not stable and will change under different environmental conditions. 
% Thus, attackers have a very low opportunity to evade \name{}, even if they know the internal design details of \name{}.  

\noindent \textbf{End-to-end Attack.}
% 2020-05-22-LFT 为了更有效的衡量我们整个检测流程的鲁棒性，设计了一个optimization-based attack来进行整个流程的end-to-end攻击。
To evaluate the robustness of our entire detection process, we build an optimization-based end-to-end attack. Because of the non-differentiable step of binarization, \name{} cannot be directly attacked with a gradient-based approach. Backward Pass Differentiable Approximation (BPDA)~\cite{athalye2018obfuscated} is an approach designed for this situation, which runs through the whole process directly to get the prediction and calculates gradients using a differentiable approximation with respect to the prediction. More specifically, we use $\hat{\beta}_{i, j}=\frac{1}{1+e^{-t\left(L_{i, j}-T\right)}}$ to approximate the binarization step, where $t$ is the temperature increasing along iterations to generate result closer to binarization, $L_{i, j}$ denotes the value at the position $(i, j)$ in the generated heatmap, and $T$ is the binarization threshold. Then, 
% instead of using $(1-\beta)*\text{image}+\beta*\text{filling\_pattern}$ as the intermediate input, 
we replace $\beta$ with $\hat{\beta}$. When generating attacks, we initialize $t$ as 1 and increase it by 1 after 100 iterations. The step size and iterations are set as 1 and 2000, respectively, using an Adam optimizer. We randomly select 200 images and run attacks targeting \emph{toaster} individually against each image, with the knowledge of all the internal details of the defense. %The FPR is set the same as above (i.e., 6\%). 
Note we ignore some physical constraints, such as location invariance and printability, to enhance the capability of attackers. Without \name{}, a total of 79 images are misclassified, but none of them are successfully attacked into the target label. With \name{}, the attack success rate is 25\%. The overall robustness of \name{} against BPDA is 75\% when 7\% of the whole image is perturbed. %It can become more robust under the physical constraints or with less white-box knowledges. 
This result largely surpasses DW~\cite{hayes2018visible} with only 5\% robustness when 5\% of the whole image is perturbed~\cite{chiang2020certified}. 
%
% Therefore, we can see that Taintradar is strongly robust even if the attacker achieve completely white-boxed 
% attack, and can effectively limit the targeted attack. When part of parameters ($k$ or $\delta R$) or filling content are unknown to attacker, the robustness could be stronger
%
% 为什么taintradar的实验结果比DW更稳定呢。我们猜测，因为在检测时，taintradar（高层卷积层）更关注区域性的内容，对像素层面的修改具有较高的鲁棒性，局部的修改像素不会过多的影响区域估计的结果。而DW则相反，像素级别的图像操作会比较容易被攻击者利用（比如生成较为稀疏的特征图），在膨胀操作后不能达到预期的密度，进而被丢弃。
%
One main reason is that \name{} leverages regional information of deep representations at intermediate layers, which is more robust against pixel-level changes than methods using pixel-prediction correspondence. Thus, bypassing \name{} restricts the ability of achieving a misclassification attack in a small and contiguous area, and vice versa.
%!TEX root=./main.tex

\section{Discussion}\label{sec:discussion}
\label{discussion}
%\subsection{Future Works}
%\noindent \textbf{GAN-Based Inert Patterns in Estimating Malicious Regions.}
\noindent \textbf{GAN for Accurate Region Estimation and Image Recovery.} 
\label{GAN-Based-inpaint}
In the current design of \name{}, we utilize filling patterns to fill the critical regions so as to restore the original information of the images. Actually, 
%In our detection module, we first remove the intersection region, then fill this region with inert pattern. As for benign images, the intersection region is confined to a small area, 
we can leverage Generative Adversarial Networks (GAN) to inpaint this region instead of filling it with a filling pattern. Thus, we can restore the original information with a reduced FPR.  We leave it as an interesting future work. 
%. We can further select smaller parameter $K$ and $\Delta R$ to improve the detection rate against adversarial images. In case we cannot find the attacked region accurately, there may exist residual noise in the images. At this time, we are required to 
%design a better mechanism to improve the performance of GAN-based inpainting. This is our further work.
%LFT-2020-01-17 在检测的过程中，我们使用了inert pattern的填充方法来对图像的移除区域进行填充，获得了较好的对恶意样本的区分效果。
%LFT-2020-01-17 由于benign样本在求反向图交集之后得到的区域较小，我们考虑可以在这一步骤中引入GAN来对这一区域进行填充，这样的方法可以进一步降低False Positive。但对于对抗样本，在有剩余噪音的情况下用GAN恢复的行为不可预知，该方法需要进一步的工作验证
%\noindent \textbf{Improving Image Recovery for Accurate Classification.}
%In many scenarios such as autopilot and security monitor, it is necessary to get the correct classification outputs even if attacked. This requires us not only to detect adversarial images, but also recover effectively to get the correct classification results.
Moreover, it is desirable to have more accurate image recovery in some scenarios, e.g., autopilots. %to 
%achieve this, 
Similarly, we can utilize GAN to recover the original label by leveraging image characteristics, such as image gradient, to eliminate the impact of residual perturbations. 
\noindent \textbf{Defenses against Backdoor Attacks.}
%LFT-2020-01-17 与adversarial attack的post-training不同，Backdoor attack是一种在模型训练时发起的常见的针对攻击（引用相关工作）。
%LFT-2020-01-17 它不在我们的attack model范围之内，但是
In this paper, we focus on localized adversarial examples without tampering with the model. Unlike post-training processes in the adversarial examples, the backdoor attack is a targeted attack launched during model training. Gu et al.~\cite{gu2017badnets} first demonstrate that CNN models can be easily backdoored by injecting poisoned data in the training dataset. Moreover, Liu et al.~\cite{Liu2018Trojaning} show that they can launch trojan attacks by modifying some specific neurons without access to the training dataset. %Trojaned models perform well on benign images, but behave badly when the backdoor trigger appears.
By using preliminary experiments, we find that the final outputs of \name{} on images with backdoor triggers are also centralized on the attacked area.
% and the adversarial patches under our approach are very similar. They both construct attacks by altering the activation distribution in the model based on localized patterns. 
Thus, \name{} can potentially be applied to defend against this type of attack.

% \noindent \textbf{Limitation of \name{}.} \name{} is rooted on the last convolutional layer, whose scale is usually smaller than the input image, e.g., 14$\times$14 for VGG16, 16X smaller than the input size of 224$\times$224. In this case, the desired 

% when the perturbed areas are scattered in the image each with a granularity smaller than 16 pixels, these areas will have less chance being located by \name{}.

\noindent \textbf{Defenses against Attacks on Traffic Signs.}
The traffic sign attack~\cite{eykholt2018robust} is another stealthy attack in the physical world, which uses stickers or camouflage graffiti to fool neural networks. For example, an attacker can put some stickers on a stop sign, which will not be perceived by human eyes but can cause misclassification to
the classifier.  %Although such malicious stickers that are constructed by performing the back-propagation  
%are similar to adversarial patches, the training 
However, The dimensions of the input features in road signs are very small in the commonly used models, e.g.,  
road sign images are normally resized to 32$\times$32 before the model is trained~\cite{sermanet2011traffic,eykholt2018robust}. 
Thus, the original attacked regions become a number of pixels in the resized image. Since our critical region estimation algorithm leverages deep representations at intermediate layers and neglects the input-prediction correspondence, it cannot locate pixels in images when they are scattered. In other words, as the granularity of the last convolutional layer is significantly larger than each scattered perturbed area, from the perspective of our approach, the attack stops to be localized. We consider addressing this problem using the traditional pixel-level threat model in our future work.

\section{Related Work}
\label{sec:background}

%Prior arts related to the paper fall into three categories:

%In this section, we first review some typical adversarial attacks and trojan backdoor attacks respectively, then introduce the related defense mechanisms and compare ours with some closest work.

\noindent \textbf{Adversarial Examples against Neutral Networks. }
Recent studies
\cite{szegedy2013intriguing, moosavi2016deepfool, moosavi2017universal,dong2019evading,carlini2017towards, su2019one, goodfellow2014explaining,papernot2016limitations} show that neural networks can be easily fooled by adversarial examples. They generate imperceptible perturbations bounded by a norm-ball constraint, e.g., $l_0$, $l_2$, or  $l_{\infty}$, in benign images so as to create traditional pixel-level adversarial examples. 
%Since the discovery of adversarial examples
%, many approaches to construct adversarial examples have been proposed by adding imperceptible perturbation
%\cite{
Normally, these attacks cannot be physically implemented in a real world environment. 
%hard to work in the real world. 
In this paper, we focus on localized attacks that can build small and physically visible \patch{}.
For example, adversarial patch can be printed and placed in an image to deceive the classifier~\cite{brown2017adversarial}, and traffic road signs with inconspicuous stickers might be misclassified by self-driving cars~\cite{eykholt2018robust}.  
Moreover, state-of-the-art face recognition systems can be fooled by \patch{}, e.g., crafted glasses~\cite{sharif2016accessorize} and graffiti stickers on hats~\cite{komkov2019advhat}.

\noindent \textbf{Defenses against Traditional Adversarial Examples.}  
Most existing defenses focus on imperceptible adversarial perturbations~\cite{papernot2016distillation, cai2018curriculum, song2017pixeldefend, kurakin2016adversarial,ghosh2019resisting, madry2017towards, gowal2018effectiveness, mirman2018differentiable}. Papernot et al.~\cite{papernot2016distillation} proposed defensive distillation extracting the key information from a pretrained DNN to improve the resilience of a model to adversarial examples. Also, an autoencoder (AE) can be applied as a denoiser to purify adversarial effects of an input~\cite{ghosh2019resisting}. Similarly, PixelDefend~\cite{song2017pixeldefend} projects the adversarial input back to the training distribution. Moreover, adversarial training~\cite{kurakin2016adversarial, madry2017towards, cai2018curriculum} improves the robustness of the DNN models by training against known attacks. Certified robustness approaches~\cite{gowal2018effectiveness, mirman2018differentiable} give a lower-bound on the adversarial accuracy. 
However, these defenses cannot be directly applied to throttle localized adversarial examples that are visible and constrained to small regions.

\noindent \textbf{Defenses against Localized Adversarial Examples.}  
%We focus on defense solutions against localized attacks, while there 
Some defense approaches\cite{ chiang2020certified, chou2018sentinet, hayes2018visible, naseer2019local, wu2019defending} have been proposed to resist localized adversarial examples.
%However, they cannot handle localized attacks. 
%
Input transformation approaches~\cite{naseer2019local, hayes2018visible} focus on mitigating the effect of adversarial perturbations and recovering to the original label of the benign input.
% Naseer et al.~\cite{naseer2019local} introduced localized gradient smoothing (LGS) based on the observation that pixel values tend to change sharper in \patch{} than in benign objects. Hayes et al.~\cite{hayes2018visible} proposed using the density of large gradient pixels to locate the adversarial object and replace them by averaged nearby pixels. 
However, none of these approaches show robustness under the BPDA attack~\cite{athalye2018obfuscated,chiang2020certified}. 
Chiang et al.~\cite{chiang2020certified} transferred certified robustness~\cite{gowal2018effectiveness, mirman2018differentiable}, which gives a certificate when an output lies in the interval bound formed during the training process. Wu et al.~\cite{wu2019defending} combined a new abstract attack model that represents physically realizable attacks with adversarial training~\cite{madry2017towards, cai2018curriculum} to increase the robustness of a model. However, both robustness approaches require extra training 
with high training overhead and cannot work well at ImageNet scale~\cite{chiang2020certified, kurakin2016adversarial}. Also, robustness approaches do not block out adversarial examples completely and cannot be used to guard existing models. Thus, a robust detection approach is in urgent need. SentiNet~\cite{chou2018sentinet} is a detection approach that sits the closest to our threat model. It leverages the property that adversarial objects can generalize to a large distribution of inputs; however, it is not held by localized partial attacks that only target a small subset of source labels, e.g., eye-glasses attack~\cite{sharif2016accessorize}. Moreover, the expensive selective search and testing algorithm make it hard to be deployed in time-bounded systems. % We compare it with \name{} in Section \ref{sec:Experiments}.

\section{Conclusion}
In this work, we propose TaintRadar to defend against localized adversarial examples. TaintRadar is the first generic  defense system that can detect various types of localized adversarial examples, in particular the partial attacks such as the eye-glasses attack. % compared with other existing methods. 
TaintRadar uses feature maps in the last convolutional layers to identify critical regions in images, and accurately identifies the differences between benign and malicious images by evaluating the impacts of the regions. % we analyze the root cause of limited defense capabilities for Grad-CAM based on the principle of localized adversarial attack. Therefore, TaintRadar is proposed and proved capable of detecting attacks effectively. 
We evaluate the effectiveness and the robustness of \name{} using different settings in two typical scenarios, i.e., scene classification and face recognition. Experimental results show that \name{} can detect partial attacks, which cannot be captured by all existing defenses. %To further demonstrate that TaintRadar can perform well in the real physical world, we simulate the attacks in a laboratory environment, 
Moreover, we show that \name{} can effectively throttle the localized attacks in the real world. %be deployed in the physical world. In addition, to fully evaluate the defense capabilities, we also demonstrate the robustness of TaintRadar from aspect of adaptive attacks and attack variants including different generalization capabilities, sizes and shapes to fully evaluate TaintRadar. More importantly, our approach can be run in a real-time environment and is convenient to deploy.

\section*{Acknowledgements}
This work is supported in part by NSFC under Grant 61572278, BNRist under Grant BNR2020RC01013 and U.S. ARO under Grant W911NF-17-1-0447. Qi Li is the corresponding author of this paper.

% \clearpage
%-------------------------------------------------------------------------------
\bibliographystyle{IEEEtran}
\bibliography{ref}

%\clearpage
%\input{Appendix}

\end{document}